\documentclass{article}

\usepackage{microtype}
\usepackage{graphicx}
\usepackage{tikz}
\usetikzlibrary{fit}
\usetikzlibrary{positioning}
\usepackage{booktabs} 

\usepackage{hyperref}
\usepackage{adjustbox}
\usepackage{amssymb}
\usepackage{amsmath}
\usepackage{caption}
\usepackage{subcaption}

\usepackage{dblfloatfix}
\usepackage{cleveref}
\usepackage{multirow}

\newcommand{\imageadv}{$\mathbf{I}_{\text{pert}}$}
\newcommand{\imagemal}{$\mathbf{I}_{\text{mark}}$}

\usepackage[accepted]{icml2021}

\icmltitlerunning{Markpainting: Adversarial Machine Learning meets Inpainting}

\begin{document}
\twocolumn[{%
\renewcommand\twocolumn[1][]{#1}%
\icmltitle{Markpainting: Adversarial Machine Learning meets Inpainting}



\icmlsetsymbol{equal}{*}

\begin{icmlauthorlist}
\icmlauthor{David Khachaturov}{equal,to}
\icmlauthor{Ilia Shumailov}{equal,to,goo}
\icmlauthor{Yiren Zhao}{to}
\icmlauthor{Nicolas Papernot}{goo}
\icmlauthor{Ross Anderson}{to}
\end{icmlauthorlist}

\icmlaffiliation{to}{Computer Laboratory, University of Cambridge}
\icmlaffiliation{goo}{University of Toronto and Vector Institute}

\icmlcorrespondingauthor{Ilia Shumailov}{ilia.shumailov@cl.cam.ac.uk}

\icmlkeywords{Machine Learning, ICML, Inpainting, Adversarial ML, Security}

\usetikzlibrary{positioning,fit,calc}
\tikzset{
	block/.style={draw,thick,text width=1.5cm,minimum height=1cm,align=center},
	line/.style={-latex}
}
\vskip 0.3in
}]



\printAffiliationsAndNotice{\icmlEqualContribution} 

\begin{abstract}

Inpainting is a learned interpolation technique that is based on generative modeling and used to populate masked or missing pieces in an image; it has wide applications in picture editing and retouching. Recently, inpainting started being used for watermark removal, raising concerns. In this paper we study how to manipulate it using our \emph{markpainting} technique. First, we show how an image owner with access to an inpainting model can augment their image in such a way that any attempt to edit it using that model will add arbitrary visible information. We find that we can target multiple different models simultaneously with our technique. This can be designed to reconstitute a watermark if the editor had been trying to remove it. 
Second, we show that our markpainting technique is transferable to models that have different architectures or were trained on different datasets, so watermarks created using it are difficult for adversaries to remove.
Markpainting is novel and can be used as a manipulation alarm that becomes visible in the event of inpainting. Source code is available at: \url{https://github.com/iliaishacked/markpainting}.



\end{abstract}



\section{Introduction}
Improvements to machine learning (ML) have enabled automatic content creation~\cite{unpublished2021dalle} and manipulation~\cite{yu2018generative}: a user just needs to provide an image and describe the changes they want as the input to a generative model~\cite{goodfellow2016nips,korshunova2017fast,DeOldify}. Computer graphics tools brought us digital \emph{inpainting}: programs such as Photoshop enable manipulation of digital images with powerful software and, more recently, ML support~\cite{photoshopml}. Modern inpainting software lets the user select a patch to be filled in; it then fills this area in with artificially generated content.

One increasingly popular application of inpainting is the removal of objects from photographs. This can be done for malicious purposes. For example, many images are distributed with a watermark that asserts copyright or carries a marketing message; people wishing to reuse the image without permission may want to remove the mark and restore a plausible background in its place. This naturally leads to the question of how we can make watermarks more robust, i.e. difficult to remove. There is substantial literature on using classic signal-processing techniques for mark removal, e.g. from~\citet{CMBF2007}, but such tricks predate recent advances in ML and inpainting more specifically. 





\usetikzlibrary{positioning,fit,calc}
\tikzset{
	block/.style={draw,thick,text width=1.5cm,minimum height=1cm,align=center},
	line/.style={-latex}
}

\begin{figure*}[!ht]
    \centering
    \begin{subfigure}[t]{.425\textwidth}
        \centering
        \resizebox{\textwidth}{!}{%
        \begin{tikzpicture}
            \node[inner sep=0pt] (adversarial)
                {\includegraphics[width=2.5cm]{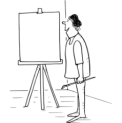}};
            \node[inner sep=0pt, right=0.5cm of adversarial] (perturbation)
                {\includegraphics[width=2.5cm]{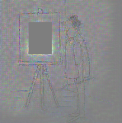}};
            \node (left-paren) [left = 0cm of adversarial] {$\left(\vphantom{\includegraphics[width=.1\textwidth]{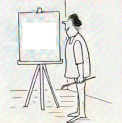}}\right.$};
            \node (right-paren) [right = 0cm of perturbation] {$\left.\vphantom{\includegraphics[width=.1\textwidth]{images/intro_explanation/adversarial.png}}\right)$};
            \node[draw,dotted,fit= (left-paren) (adversarial) (perturbation) (right-paren), inner sep=0.25cm] (input) {};
            
            \node[block, below left = 1cm and -2.5cm of input] (model) {Infill\\Model};
            
            \node[inner sep=0pt, above left = -0.25cm and 0.75cm of model] (mask_p)
                {\includegraphics[width=1cm]{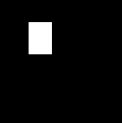}};
            \node[draw,dotted,fit=(mask_p)] (mask) {};
            
            \node[inner sep=0pt, below=0.5cm of model] (result)
                {\includegraphics[width=2.5cm]{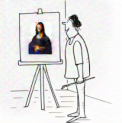}};
            \node[draw,dotted,fit=(result)] (output) {};
            \node[inner sep=0pt, right=1cm of output] (target_img)
                {\includegraphics[width=2.5cm]{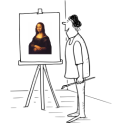}};
            \node[draw,dotted,fit=(target_img)] (target) {};
            \node[block, above = 0.75cm of target] (loss) {$\mathcal{L_{\text{mark}}}$\\Loss};
            
            \path (adversarial.east) -- (perturbation.west) node[midway] {$+$};
            \draw[->] (input.south -| model.north) -- node[midway, right] {input image} (model.north);
            \draw[->] (mask.south) |- node[midway, below right] {mask} (model.west);
            \draw[->] (model.south) -- (output.north);
            \draw[->] (target.north) -- node[midway, right] {target} (loss.south);
            \draw[->] (output.north) ++(1.25, 0) |- node[midway, below right] {result} (loss.west);
            \draw[->] (loss.north) -- node[right] {perturbation} (loss.north |- perturbation.south);
        \end{tikzpicture}
        }
        \caption{}
        \label{fig:explanation}
    \end{subfigure}
    \begin{subfigure}[t]{.425\textwidth}
        \centering
        \includegraphics[width=\textwidth]{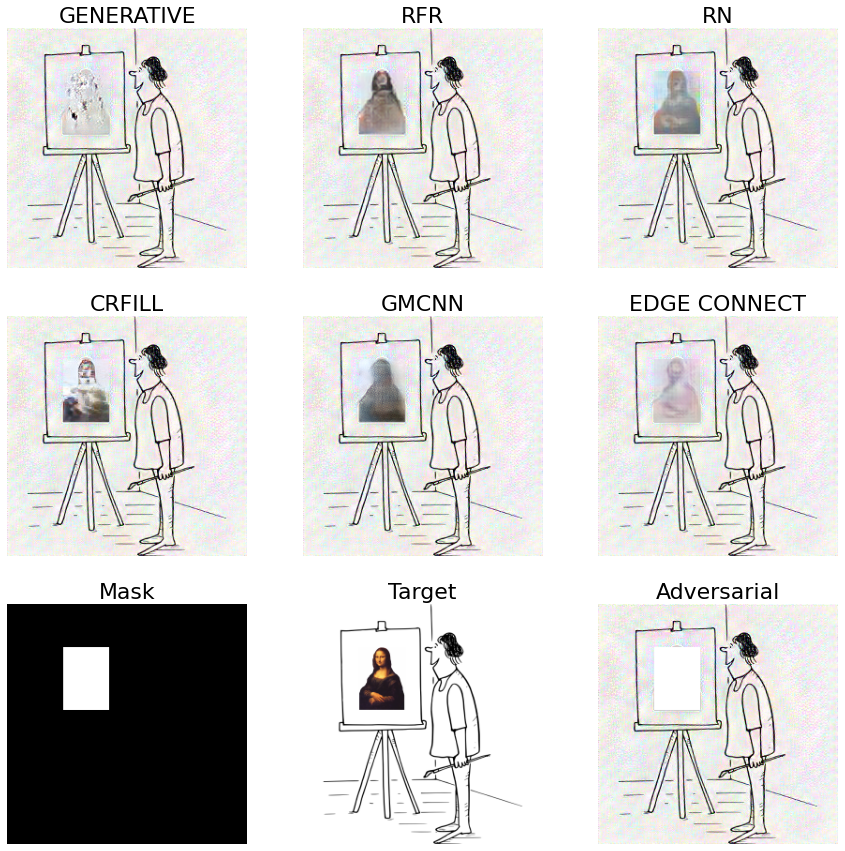}
        \caption{}\label{fig:example}
    \end{subfigure}
    \caption{Demonstration of the proposed markpainting technique. The target image is set to be Leonardo da Vinci's \emph{La Gioconda} pasted onto the otherwise-blank cartoon canvas. \Cref{fig:explanation} shows a visual abstract of the proposed markpainting technique, using the CRFILL model. \Cref{fig:example} shows the application of markpainting to multiple different inpainting models simultaneously --- our technique can target multiple models at once and is not limited to just a single model. The \emph{Adversarial} pane shows the combination of the original input image and the resulting perturbations. The top six images show the result of various inpainters filling-in the rectangular patch on the canvas as defined by the mask. Note that all six inpainters use the same input, namely \emph{Adversarial}. Original cartoon from \href{https://freesvg.org/artist-with-blank-canvas}{freesvg}.
    }
\end{figure*}

\begin{figure}[!th]
    \centering
    \includegraphics[width=\linewidth]{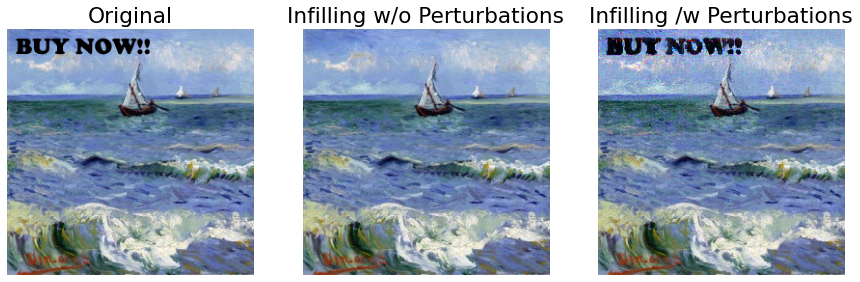}
    \caption{Example of countering watermark removal using markpainting on Vincent van Gogh's \emph{Boats at Sea}. The left-most image depicts the original image with the watermark. The middle image is the result of inpainting the mark without any perturbations, resulting is the successful removal of the watermark. The right-most image contains generated perturbations and has been treated with an inpainter for watermark removal; the output simply restores the mark. Performed on the CRFILL inpainting model with $\epsilon=0.3$.
    }
    \label{fig:watermark_crfill}
\end{figure}

In this paper we investigate whether ML inpainters can be manipulated using techniques adapted from the field of adversarial machine learning. Our technique, which we dub \emph{markpainting}, allows for arbitrary manipulation of how inpainters fill in the patch defined by a given 2-bit image \emph{mask}. We do this by setting an arbitrary target image which we wish to appear in the filled-in area. We then generate \emph{perturbations} -- small pixel-wise augmentations -- which are applied to the original image to manipulate the inpainting algorithms into producing something resembling our target. For example, in~\Cref{fig:explanation}, the original image is a black-and-white cartoon; we set the target image to be the same cartoon but with \emph{La Gioconda} pasted onto the otherwise blank canvas. After the application of our technique, the perturbations to the original image ensure that the resulting infilled patch does indeed resemble our target.

We find that the introduction of minor perturbations to input images can force many inpainters to generate arbitrary patches --- even patterns not present in the training datasets. Consequently, setting the target to be the original image and applying our markpainting technique makes the image robust against watermark removal as shown in~\Cref{fig:watermark_crfill}. The original (left-most) image has an unsightly watermark that was removed successfully in the middle image by an inpainter. However, after treating the original image with our markpainting technique -- setting the target to be the original image itself, to preserve the watermark -- the attempt to paint out the watermark fails.


\Cref{fig:example} demonstrates the effect of markpainting on six different inpainters. The resulting markpainted sample (bottom right in~\Cref{fig:example}) is a combination of the original image (top left in~\Cref{fig:explanation}) and the accumulated perturbations. We can see that \emph{La Gioconda} (the target) appears on the canvases (the patch to fill in as dictated by the mask) of the final inpainted images (top two rows in~\Cref{fig:example}). These final images are obtained by running the markpainted sample through each of the inpainting models.

We find that markpainting can work even if the colors and structures of the target image are not present in the input image itself or the dataset the model was trained on. We evaluate the extent to which markpainting transfers from one inpainter to another and within the same inpainter trained on different datasets; the impact of perturbation size; and the viability of mask-agnostic markpainting. 

Overall, we make the following contributions: 
\begin{itemize}
    \item We show that inpainting can be manipulated to produce arbitrary content, a technique we name \textit{markpainting}.
    \item We present a mask-agnostic markpainting method that works regardless of the mask used.
    \item We evaluate the performance of markpainting thoroughly and find that markpainting a specific target is significantly more effective against more advanced inpainters (a $38\%$ reduction in loss to target in the case of a weak Generative model, compared to a $78\%$ reduction in EdgeConnect's case).
    \item In a robustness test, we show that markpainted samples sometimes transfer within the same inpainter trained on different datasets, and across different inpainters for markpainting with a target. 
\end{itemize}

\section{Broader Impact and Motivation}

\begin{figure}[h!]
    \centering
    \includegraphics[width=\linewidth]{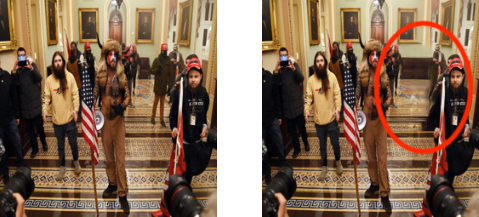}
    \caption{Photo taken from the 6 January 2021 raid on United States Congress. Original on the left; the right photo has been modified using an inpainter to remove a participant. It is near impossible to tell which of the two images is the original one, without closer inspection.}\label{fig:congress}
\end{figure}

Malicious actors now manipulate public discourse with artificially generated or manipulated images, such as deepfakes~\cite{goodfellow2014generative,zhang2020deepfake}. For example, as shown in~\Cref{fig:congress}, it takes no special knowledge to remove a participant from a photo of the 6 January 2021 raid on United States Congress; this is not noticeable without inspecting the image closely. This motivating example led us to study the capacity of inpainting tools to remove or replace objects in images.

Markpainting can provide protection against evidence tampering by preserving the integrity of published images. Consider an image of a crowd and an attacker who wants to forge evidence by removing a person from the crowd. The defender -- e.g.\ the distributor of the image -- does not know which person will be removed, but wants to stop the attacker. If they use our mask-agnostic markpainting technique with a solid color target image (such as pure red), then any attempt to remove a person from the image via inpainting will result in a red patch, clearly marking the image. In practice one would use more subtle techniques, which we discuss later.






\section{Related Work}

Humans have been restoring paintings for centuries. As ultraviolet light degrades both pigment and bindings in paint, exterior paintwork needs regular reworking; and although artworks kept indoors deteriorate more slowly, they still require upkeep from time to time. Images are touched up for other reasons; after Trotsky fell from favor in Russia, he was airbrushed out of numerous paintings. Digital inpainting is newer, going back to the 1990s when computer-graphics tools started to become both capable and widespread. Early approaches included patch search methods~\cite{bertalmio2000image, osher2005iterative} and texture synthesis~\cite{efros1999texture,barnes2009patchmatch}. Those approaches can only work with small missing regions because of the lack of semantic understanding; they are usually computationally expensive because of the time taken to find close matches to missing objects in a large corpus of data~\cite{hays2007scenecompletion}. 

Recent advances in machine learning have enabled more semantically-aware inpainting. In 2016, \citet{pathak2016contextencoders} presented Context Encoders -- CNNs trained to predict the contents of an arbitrary image based on its surroundings. They used L2 and an adversarial loss as in generative adversarial networks~\cite{goodfellow2014generative}. 
In 2017, \citet{iizuka2017globally} built on this work by splitting the discriminator into three: a completion network, a local discriminator and a global discriminator. This architecture allowed inpainting of images of arbitrary size and helped maintain local consistency. Poisson blending allowed further refinement and sharpening of the image. 
In 2018, \citet{wang2018image} proposed a Generative Multi-column Convolutional Neural Network (GMCNN) with three sub-networks: a generator to inpaint the image, global and local discriminators and a pretrained VGG network for Implicit Diversified Markov Random Fields (ID-MRF) loss calculation. They use filters of different sizes to capture information at different granularity levels, which allowed more fine-grained inpainting.
In 2019, \citet{nazeri2019edgeconnect} used image structure knowledge and developed a two-stage model composed of an edge generator and an image generator.\vspace{0.5\baselineskip}

Recently, there has been significant work in this field. \citet{li2020recurrentfeature} proposed Recurrent Feature Reasoning (RFR), an inpainting method based on Knowledge Consistent Attention modules. RFR recurrently infers the hole boundaries, then uses them to solve more complex parts of the image. It is split into three parts: an area identification model, a feature reasoning module and a feature-merging operator designed to combine intermediate feature maps. The networks are trained to optimize VGG perceptual and style losses.  
\citet{li2020deepgin} proposed a deep generative inpainting network named DeepGin, using a customized ResNet block \cite{he2016deep} to allow different receptive fields so that information from both local and distant regions can be gathered efficiently.
\citet{jie2020inpainting} built on Nazeri's work with a shared generator to generate both the completed image and its corresponding structures, placing the inpainting problem into a multi-task learning setup.
\citet{yu2020region} investigated the feature normalization problem in the context of image inpainting, and proposed a spatially region-wise normalization for image inpainting.
\citet{zeng2020image} proposed using a contextually-aware reconstruction loss to replace the contextual attention layers so a network could explicitly borrow from a known region as a reference to inpaint images.





On the adversarial machine learning side of things, in 2013 two separate teams led by Szegedy and Biggio discovered adversarial examples which, during inference, cause a model to output a surprisingly incorrect result~\cite{szegedy2013intriguing,biggio2013evasion}. In a white-box environment -- where the adversary has direct access to the model -- such examples can be found using various gradient-based methods that typically aim to maximize the loss function under a series of constraints~\cite{biggio2013evasion,szegedy2013intriguing,goodfellow2015explaining,madry2019deep}. In a black-box setting, the adversary can transfer adversarial examples from another model~\cite{papernot2017practical} or approximate gradients by observing output labels and confidence~\cite{chen2017zoo}. In their various forms, adversarial examples can affect the \textit{confidentiality}, \textit{integrity} and \textit{availability} of machine learning systems~\cite{biggio2018wild,papernot2016towards,shumailov2020sponge}.

\begin{figure}[t]
    \begin{subfigure}[b]{0.6\linewidth}
        \centering
        \includegraphics[width=\linewidth]{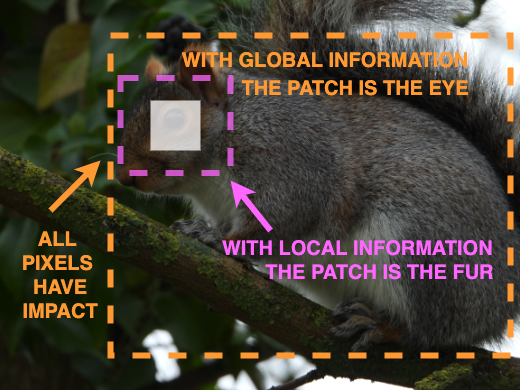}
    \end{subfigure}%
    \begin{subfigure}[b]{0.45\linewidth}
        \centering
        \includegraphics[width=\linewidth]{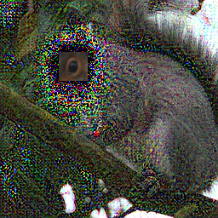}
    \end{subfigure}
    \caption{Inpainting of the squirrel eye requires both local and global knowledge. With just local knowledge only the fur patterns could be produced. Image on the right features exaggerated normalized gradients of EdgeConnect~\cite{nazeri2019edgeconnect} during the first algorithm iteration.}
    \label{fig:sqexample}
\end{figure}





\section{Methodology}

\subsection{Inpainting}

\textit{Inpainting} fills in information that is missing in an input image. During training, a part of the image is masked out and the inpainter aims to learn how to restore this area. 

We define an input RGB image $\mathbf{I} \in \mathbb{R}^{H \times W \times 3}$ and a binary mask $\mathbf{M}\in \mathbb{R}^{H\times W}$. The binary mask $\mathbf{M}$ has $0$s for the areas to be inpainted and $1$s otherwise. We then assume an inpainter $f$, that populates the region covered by $1-\mathbf{M}$ taking as input masked input $\hat{\mathbf{I}} = \mathbf{I} \odot (1-\mathbf{M})$, where $\odot$ represents the Hadamard product. The function $f$ was trained to minimize dissimilarity $\mathcal{L_{\text{train}}}$ between $\hat{\mathbf{I}}$ and $\mathbf{I}$. Training here may involve images of different sizes and irregular masks depending on the system. 

\subsection{Markpainting} 
\label{sec:method:markpaint}

We present two different flavors of markpainting: targeted and mask-agnostic. Targeted markpainting forces the reconstruction to resemble the target image, whilst mask-agnostic markpainting aims to generalize the technique to work with an arbitrary mask. These are presented in~\Cref{alg:attackalg} and~\Cref{alg:eotalg} respectively. \Cref{alg:attackalg} is visualized in~\Cref{fig:explanation}. 
The formal setup is similar to adversarial example generation~\cite{szegedy2013intriguing, madry2019deep}, where the perturbation $\eta$ is accumulated iteratively from scaled gradients ($\epsilon'\text{sign}(\nabla_\mathbf{I}\mathcal{L_{\text{mark}}}(\mathbf{\theta}, \mathbf{I}, \mathbf{T}))$).

We define $\mathcal{L_{\text{mark}}}(\theta, x, x') = \mathcal{L_{\text{network}}}(\theta, x) + \alpha l_2(x-x')$, where $\mathcal{L_{\text{network}}}$ is the VGG perceptual loss~\cite{johnson2016perceptual} and $l_2$ is MSE loss. We use the VGG perceptual loss to measure human visual similarity of markpainting, which is usually missed by pure L2 loss. L2 penalizes large deviations from the target, whilst VGG promotes human-understandable granularity. We set $\alpha=4$, based on experimentation. The effect of different $\alpha$ values on the markpainted result can be found in Section D of our Appendix. 

Notice that the perturbation propagated to the natural input is $(\eta \odot (1-\mathbf{M}))$, because the regions to be infilled are masked out and do not receive gradients.

The technique aims to find a perturbation $\eta$ with a given perturbation budget $\epsilon$ such that the used dissimilarity function $\mathcal{L_{\text{mark}}}$ parameterized by $\theta$ is minimized.

\begin{equation*}
\begin{aligned}
& \underset{\eta}{\text{minimize}}
& & \mathcal{L_{\text{mark}}}(\theta, f((\mathbf{I + \eta}) \odot (1-\mathbf{M})), \hat{\mathbf{I}}) \\
& \text{subject to}
& & ||\eta||_p < \epsilon
\end{aligned}
\end{equation*}

$||\eta||_p $ is the $l_p$ norm of $\eta$ and in this paper we use $p=\infty$.

We represent the original input image using $\mathbf{I}$, the original image with our carefully crafted perturbation as \imageadv,
the naturally inpainted image using $\mathbf{I}_{\text{benign}}$, and the inpainted results of \imageadv\ as \imagemal~. We denote the target image using $\mathbf{T}$, and the mask is represented using $\mathbf{M}$.

\begin{algorithm}[tb]
   \caption{General markpainting algorithm}
   \label{alg:attackalg}
\begin{algorithmic}
   \STATE {\bfseries Input:} image $\mathbf{I}$, mask $\mathbf{M}$, target $\mathbf{T}$, perturbation step size $\epsilon'$, iterations $t$, targeted models $\mathbf{\Theta}$
   \FOR{$j=0$ {\bfseries to} $t$}
   \STATE $\eta \leftarrow \mathbf{0}$
   \FOR{$\theta \in \mathbf{\Theta}$}
   \STATE $\eta \leftarrow \eta + \epsilon'\text{sign}(\nabla_\mathbf{I}\mathcal{L_{\text{mark}}}(\mathbf{\theta}, \mathbf{I}, \mathbf{T}))$
   \ENDFOR
   \STATE $\mathbf{I} \leftarrow \mathbf{I} - (\eta \odot (1-\mathbf{M}))$
   \ENDFOR
   \STATE $\mathbf{I}_{\text{adv}} \leftarrow \mathbf{I}$
   \STATE {\bfseries Output:} markpainted sample $\mathbf{I}_{\text{adv}}$ (combination of original input image $\mathbf{I}$ and the accumulated perturbations)
\end{algorithmic}
\end{algorithm}

We find that we can apply our technique to a collection of models $\mathbf{\Theta}$ simultaneously using a single input image $\mathbf{I}$ as detailed in \Cref{alg:attackalg}. An example result of application of markpainting to multiple models simultaneously is presented in~\Cref{fig:example} and in Section A of our Appendix, where the \emph{same} markpainted sample produces a visually-recognizable face, similar to the target, after being run through six different inpainters. 

\subsection{Mask-agnostic Markpainting}

Although~\Cref{alg:attackalg} works well against a known mask $\mathbf{M}$, there are other cases where we do not know which parts of an image might be tampered with. We adapt our technique to generate an image that will cause a system to markpaint regardless of the mask used.  
This problem is related to the construction of adversarial examples that work in physical environments under different conditions of lighting and viewing angles. We therefore extend an approach first introduced by~\citet{athalye2018synthesizing} called \emph{Expectation over Transformation} (EoT).

This extension is presented in~\Cref{alg:eotalg}. For this technique, a set of random masks is produced with a given size range $[m_\text{min}, m_\text{max}]$. We iteratively sample a single mask from the set and apply an algorithm similar to~\Cref{alg:attackalg}. We find that further adding stochasticity helps to transfer to unseen masks: we weight the gradient step with a random uniformly-distributed vector $\mathbf{U}(0, 1)$.

\begin{algorithm}[tb]
   \caption{EoT markpainting algorithm}
   \label{alg:eotalg}
\begin{algorithmic}
   \STATE {\bfseries Input:} image $\mathbf{I}$, target $\mathbf{T}$, number of masks $n$, mask size range $[m_\text{min}, m_\text{max}]$, perturbation step size $\epsilon'$, iterations $t$, targeted models $\mathbf{\Theta}$
   \STATE Initialize set $\hat{\mathbf{M}}$ to contain $n$ random rectangular masks of size $s \in [m_\text{min}, m_\text{max}]$
   \STATE Initialize $\mathbf{M} \leftarrow \varnothing$
   \FOR{$j=0$ {\bfseries to} $t$}
   \STATE $\mathbf{M} \leftarrow \hat{\mathbf{M}}_i$ for a random $0 \leq i < n$
   \STATE $\eta \leftarrow \mathbf{0}$
   \FOR{$\theta \in \mathbf{\Theta}$}
   \STATE $\eta \leftarrow \eta + \epsilon' \mathbf{U}(0,1)\text{sign}(\nabla_\mathbf{I}\mathcal{L_{\text{mark}}}(\mathbf{\theta}, \mathbf{I}, \mathbf{T}))$
   \ENDFOR
   \STATE $\mathbf{I} \leftarrow \mathbf{I} - (\eta \odot \mathbf{M})$
   \ENDFOR
   \STATE $\mathbf{I}_{\text{adv}} \leftarrow \mathbf{I}$
   \STATE {\bfseries Output:} markpainted sample $\mathbf{I}_{\text{adv}}$ (combination of original input image $\mathbf{I}$ and the accumulated perturbations)
\end{algorithmic}
\end{algorithm}

\subsection{Why does it work?}

Inpainting is a complex task, with neural networks trained to manipulate images of arbitrary size and with arbitrary patches. Furthermore, modern inpainters can fill irregular holes. As they are trying to be semantically aware and display both local and global consistency, they need to understand the global scenery well. That in turn makes them dependent not only on the area around the patch, but on the whole image. Imagine trying to fill in a hole around the squirrel eye depicted in~\Cref{fig:sqexample}. Here, local information (shown in pink) would suggest that it has to be filled with fur. Global information (shown in orange) on the other hand, should tell the inpainter that the picture features a squirrel in a particular pose and that an eye should be located there. As illustrated in the gradient visualization in \Cref{fig:sqexample}, gradients focus on both the area around the eye and the rest of the image. This dependency on global information makes inpainting both complex and prone to manipulation. The markpainter does not need to concentrate their perturbation around the patch area but can scatter it all over the image. 

While at first glance markpainting seems similar to older techniques, such as ones proposed by~\citet{levincolor2004}, there are fundamental differences in the two approaches. Inpainting requires a semantic understanding of the scenery and heavily depends on global information, as shown in~\Cref{fig:sqexample}. Furthermore, markpainting can produce artifacts that are semantically meaningless for the model and not present in its training distribution. 
\begin{figure*}[t]
    \centering
    \includegraphics[width=\textwidth]{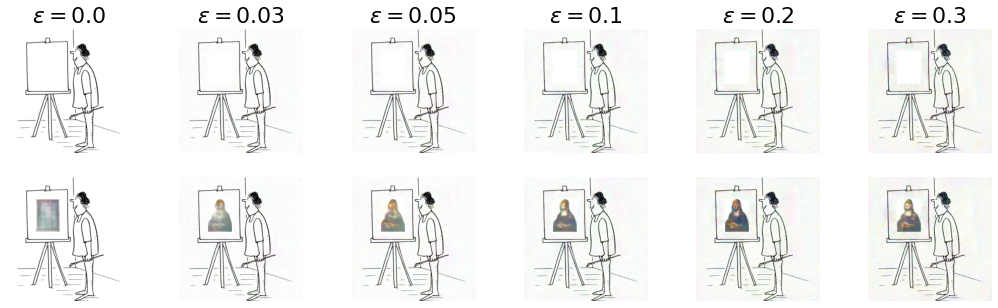}
    \caption{Inpainting with an increasing perturbation budget. Top row is the perturbed images generated using markpainting, and second row is the inpainted results of these perturbed images. We target the RN inpainter with 500 iterations and a step size of $\epsilon/100$. Note that this example is really hard, because we are filling a black and white image with color. Details are discussed in \Cref{sec:eval:targeted}.}
    \label{fig:eps_painter}
\end{figure*}

\section{Evaluation}
In this section, we evaluate the performance of targeted markpainting in \Cref{sec:eval:targeted}, and the effect of different masks and target images in~\Cref{sec:eval:impact}. \Cref{sec:eval:transfer} focuses on the transferability of the generated samples, while~\Cref{sec:eval:agnostic} discusses mask-agnostic markpainting.

\subsection{Datasets and Models}

In~\Cref{tab:datasets} we list the inpainter systems used in the evaluation. Our evaluation covers systems that provide different levels of granularity of the inpainted regions and different levels of representation. We intentionally chose a variety of inpainters with differing levels of performance and from different years. We use pretrained models provided by the authors of the respective systems. 
\Cref{tab:datasets} also indicates the datasets with which these inpainters are pretrained. A maximum perturbation budget of $\epsilon$ was used with a step size of $\epsilon' = \frac{\epsilon}{50}$ unless specified otherwise. We justify the parameter choices in Section D of the Appendix. We clip the markpainted image at each iteration to make sure that the total perturbation budget does not exceed $\epsilon$.

\begin{table}[!ht]
    \caption{Inpainters used in the evaluation }
    \footnotesize
    \centering
    \adjustbox{max width=\linewidth}{%
    \begin{tabular}{lrl}
    \toprule
    \textbf{System}  & & \textbf{Dataset} \\
    \midrule
    Generative & \cite{yu2018generative} & ImageNet~\cite{deng2009imagenet} \\
    \midrule
    GMCNN & \cite{wang2018image} & CelebA-HQ~\cite{liu2015faceattributes}\\
    \midrule
    EdgeConnect & \cite{nazeri2019edgeconnect} & \begin{tabular}[t]{@{}l@{}}Paris StreetView~\cite{doersch2012what},\\ CelebA~\cite{liu2015faceattributes},\\Places2~\cite{zhou2017places}\end{tabular}\\
    \midrule
    RFR & \cite{li2020recurrentfeature} & Paris StreetView, CelebA \\
    \midrule
    RN & \cite{yu2020region} & Places2 \\
    \midrule
    CRFILL & \cite{zeng2020image} & \begin{tabular}[t]{@{}l@{}}Places2,\\
        Salient Object Segmentation\\\cite{xiong2019foregroundaware}\end{tabular} \\
    
    \bottomrule
    \end{tabular}
    \label{tab:datasets}
    }
\end{table}


The systems are evaluated on \emph{places\_subset16}, a series of 16 randomly-selected images from the Places2 dataset~\cite{zhou2017places}\footnote{We abstain from evaluating on CelebA dataset due to ethical concerns over the dataset labels.} -- using fixed random masks of three different sizes respectively covering 5\%, 10\% and 20\% of the image. We use three solid-color targets for evaluation: pure red, green, and blue. Further details are provided in the Appendix.

\subsection{Targeted Markpainting}
\label{sec:eval:targeted}
\begin{table*}[ht]
\begin{adjustbox}{scale=0.9,center}
\begin{tabular}{@{}llrrrr|rrrr|rrrr@{}}
\toprule
\multicolumn{2}{l}{Distance to:} &\multicolumn{4}{c|}{Original} & \multicolumn{4}{c|}{Adversarial Target} & \multicolumn{4}{c}{Benign} \\
$\epsilon$ & &Loss & $l_2$ & PSNR & SSIM & Loss & $l_2$ & PSNR & SSIM & Loss & $l_2$ & PSNR & SSIM\\
\midrule
\multirow{5}{*}{$ 0.0 $}
&GENERATIVE & 0.467 & 0.111 & 12.491 & 0.250 & 0.755 & 1.186 & -0.622 & 0.036 & \textbf{0.179} & 0.000 & 134.254 & 1.000\\
&RFR & 0.279 & 0.027 & 19.156 & 0.387 & 0.433 & 0.292 & 5.464 & 0.090 & \textbf{0.001}& 0.000 & 144.880 & 1.000\\
&RN & 0.300 & 0.025 & 17.891 & 0.420 & 0.473 & 0.292 & 5.438 & 0.104 & \textbf{0.001} & 0.000 & inf & 1.000\\
&CRFILL & 0.470 & 0.109 & 13.342 & 0.319 & 0.796 & 1.205 & -0.683 & 0.044 & \textbf{0.180} & 0.000 & inf & 1.000\\
&GMCNN & 0.485 & 0.122 & 10.891 & 0.210 & 0.721 & 1.136 & -0.432 & 0.047 & \textbf{0.181} & 0.000 & inf & 1.000\\
&EDGE CONNECT & 0.290 & 0.025 & 18.198 & 0.390 & 0.422 & 0.299 & 5.383 & 0.104 & \textbf{0.001} & 0.000 & 135.133 & 1.000\\
\midrule
\multirow{5}{*}{$ 0.05 $}
&GENERATIVE & 0.441 & 0.108 & 12.121 & 0.225 & 0.623 & 1.093 & -0.281 & 0.050 & \textbf{0.289} & 0.028 & 17.483 & 0.476\\
&RFR & 0.332 & 0.040 & 15.533 & 0.325 & 0.310 & 0.232 & 6.547 & 0.139 & \textbf{0.161} & 0.019 & 18.374 & 0.630\\
&RN & 0.386 & 0.081 & 11.697 & 0.266 & \textbf{0.203} & 0.153 & 8.763 & 0.239 & 0.271 & 0.062 & 13.094 & 0.435\\
&CRFILL & 0.491 & 0.244 & 7.291 & 0.128 & 0.509 & 0.782 & 1.481 & 0.135 & \textbf{0.406} & 0.186 & 8.676 & 0.232\\
&GMCNN & 0.605 & 0.818 & 3.047 & -0.018 & 0.579 & 1.349 & -0.453 & 0.040 & \textbf{0.432} & 0.698 & 4.491 & 0.038\\
&EDGE CONNECT & 0.351 & 0.057 & 13.032 & 0.317 & 0.216 & 0.206 & 7.120 & 0.214 & \textbf{0.190} & 0.042 & 14.655 & 0.546\\
\midrule
\multirow{5}{*}{$ 0.3 $}
&GENERATIVE & 0.489 & 0.184 & 8.676 & 0.132 & 0.466 & 0.768 & 1.284 & 0.140 & \textbf{0.379} & 0.135 & 9.746 & 0.167\\
&RFR & 0.412 & 0.095 & 10.965 & 0.288 & \textbf{0.143} & 0.110 & 10.191 & 0.247 & 0.278 & 0.079 & 11.604 & 0.381\\
&RN & 0.456 & 0.154 & 8.620 & 0.216 & \textbf{0.067} & 0.049 & 14.125 & 0.322 & 0.365 & 0.140 & 8.970 & 0.315\\
&CRFILL & 0.698 & 0.896 & 0.759 & 0.035 & \textbf{0.159} & 0.060 & 14.076 & 0.695 & 0.646 & 0.884 & 0.789 & 0.047\\
&GMCNN & 0.681 & 1.112 & 0.968 & -0.067 & 0.533 & 1.457 & 0.175 & 0.081 & \textbf{0.511} & 0.993 & 1.851 & -0.066\\
&EDGE CONNECT & 0.428 & 0.112 & 10.336 & 0.285 & \textbf{0.089} & 0.091 & 11.166 & 0.268 & 0.289 & 0.103 & 10.865 & 0.398\\
\bottomrule
\end{tabular}
\end{adjustbox}
\caption{Impact of markpainting on different inpainter models.
This table reports the loss ($\mathcal{L_{\text{mark}}}$ from \Cref{sec:method:markpaint}), L2 norms, peak signal to noise ratio (PSNR) and structural index similarity (SSIM) for assessing the inpainted image quality.
Markpainting is applied to each individual inpainter and evaluated on the same inpainter with different perturbations budgets; this table is a compact version of Table 3 in our Appendix, where more epsilon values are available. In this table we highlight cases where the loss to the target image is better than to the original reconstruction. 
For these three meta-columns, we report how the markpainted patch ($\mathbf{I}_{\text{mark}} \odot \mathbf{M})$ compares to different images in different metrics.
`Original' refers to the original image patch $(\mathbf{I} \odot \mathbf{M})$, `Adversarial target' is the target image used $(\mathbf{T} \odot \mathbf{M})$, and `Benign' is the image that the model would have produced without any adversarial perturbation $(\mathbf{I}_{\text{benign}} \odot \mathbf{M})$.
Increasing the perturbation budget $\epsilon$ increases the similarity between the markpainted patch and the target but decreases similarity to the original image and benign inpainted patch.
Details are discussed in \Cref{sec:eval:targeted}.
}
\label{tab:individual_model_compact}
\end{table*}

\Cref{fig:eps_painter} illustrates the visual effect of applying markpainting to the inpainter based on Region Normalization (RN)~\cite{yu2020region} with an increasing perturbation budget. 
The top row shows the markpainted images we produced and the second row shows the final inpainted results. The inpainting task here is complex, as it requires constructing a colored patch from a black-and-white image. Even with a small budget $\epsilon=0.05$ that is barely perceptible, RN markpaints the region with a lot of detail from the target image: we can see the structure and edges of \textit{La Gioconda}. At larger $\epsilon$ values, facial details start to appear. 

\Cref{tab:individual_model_compact} presents an evaluation on the \emph{places\_subset16} dataset, averaging results from all possible input-mask-target combinations.
It is seen that larger perturbation budgets $\epsilon$ cause the dissimilarity metric $\mathcal{L_{\text{mark}}}$ and $l_2$ norm distance to reduce, and PSNR and SSIM to increase, between the markpainted image \imagemal and the target $\mathbf{T}$.
This is expected since the increased budget results in better reconstruction of the target, with the effect that the samples lose resemblance to the benign reconstruction $\mathbf{I}_{\text{benign}}$. Larger budgets help fine-grained artifacts from the target to appear in \imagemal, whilst sacrificing imperceptibility. We empirically see that $\epsilon=0.05$ is usually invisible, while allowing for a good level of reconstruction detail. 

The effectiveness of our technique when targeting multiple models simultaneously is discussed in Section A of our Appendix.


\subsection{Impact of Mask-target Choice}
\label{sec:eval:impact}
\begin{figure}[t]
    \centering
    \begin{subfigure}{.49\linewidth}
      \centering
      \includegraphics[width=\linewidth]{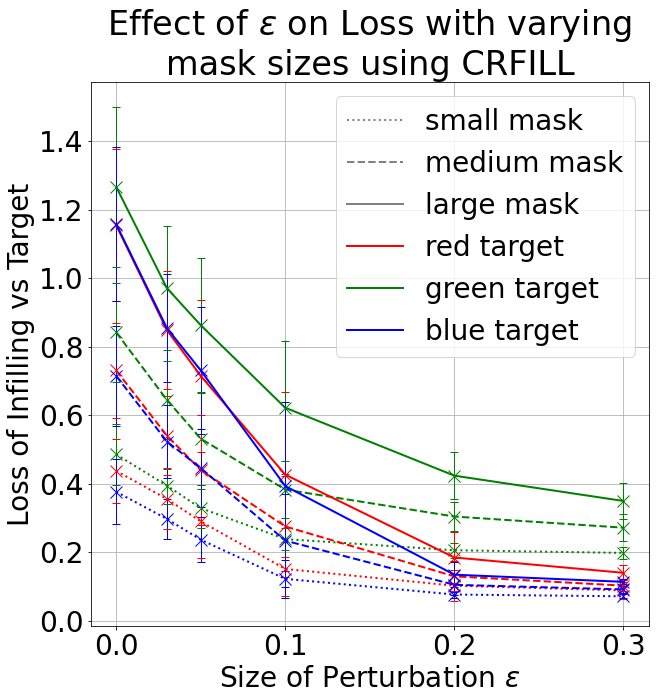}
    \end{subfigure}
    \begin{subfigure}{.49\linewidth}
      \centering
      \includegraphics[width=\linewidth]{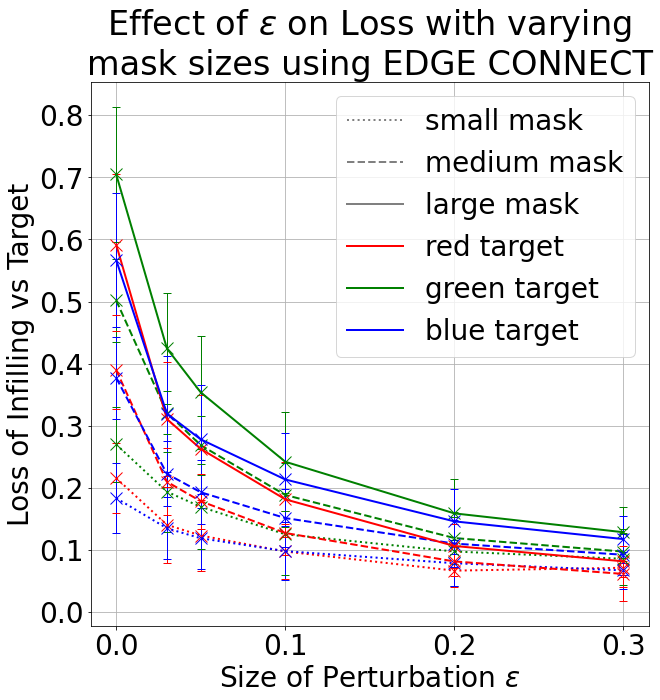}
    \end{subfigure}
    \caption{The impact that perturbation sizes have on $\mathcal{L_{\text{mark}}}$ between the markpainted patch and the target (lower is better). 
    We run markpainting over $100$ iterations. The x-axis shows different perturbation budgets and the vertical axis is the loss between \imagemal\ and $\mathbf{T}$. Results are averaged across the \emph{places\_subset16} images, with $\pm\sigma$ error bars. Details are discussed in \Cref{sec:eval:impact}.}
    \label{fig:mask_size}
\end{figure}

To illustrate the effectiveness of our markpainting technique, 
we show how it performs under varying mask sizes and target images in \Cref{fig:mask_size}.
We find that although mask size has an influence on technique performance, the color of the target image has a greater impact. 
We find that green color areas are harder to markpaint for both models, whereas blues are easiest. We suspect this is due to the source images having little green, and the training datasets perhaps also lacking this color.


\begin{figure}[ht]
    \centering
    \begin{subfigure}{.49\linewidth}
      \centering
      \includegraphics[width=\linewidth]{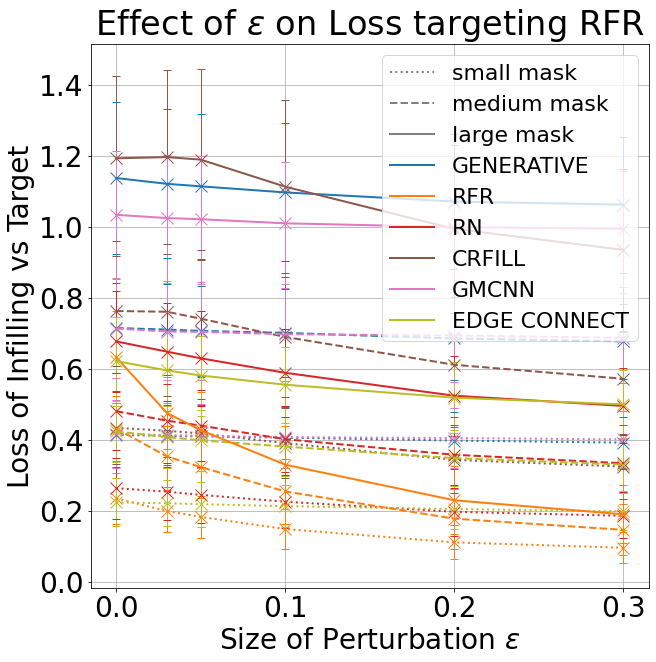}
    \end{subfigure}
    \begin{subfigure}{.49\linewidth}
      \centering
      \includegraphics[width=\linewidth]{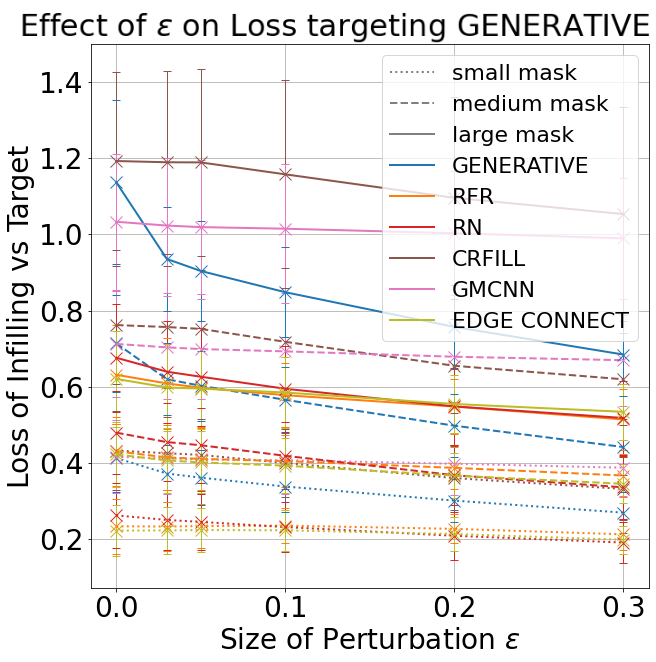}
    \end{subfigure}
    \caption{Showing technique transferability between different models using $\mathcal{L_{\text{mark}}}$ between the markpainted patch and the target. The perturbations come from targeting the model listed in the title but the errors are shown for the 6 color-coded models. Results are averaged across all possible input-mask-target combinations, with $\pm\sigma$ error bars to highlight the standard deviation in obtained results. Details are described in \Cref{sec:eval:transfer}.}
    \label{fig:transferability}
\end{figure}
\begin{figure}[t]
    \centering
    \begin{subfigure}[t]{0.49\linewidth}
      \centering
      \includegraphics[width=\linewidth]{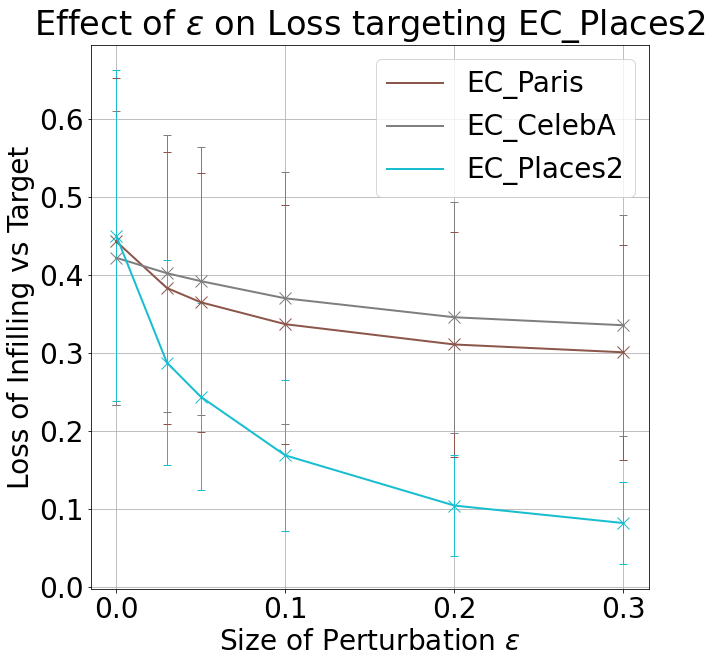}
      \caption{\scriptsize{EdgeConnect, trained on Places2.}}
    \end{subfigure}
    \begin{subfigure}[t]{0.49\linewidth}
      \centering
      \includegraphics[width=\linewidth]{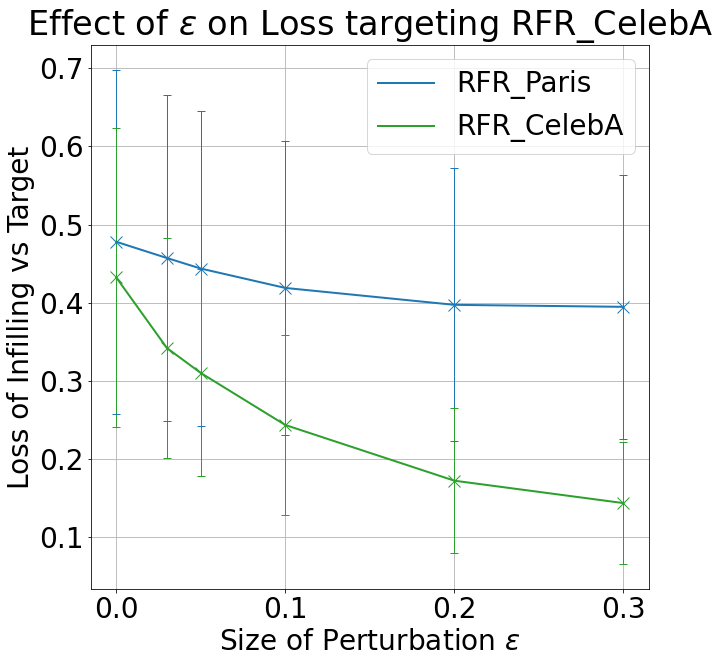}
      \caption{\scriptsize{RFR, trained on CelebA.}}
    \end{subfigure}
    \caption{Showing technique transferability between the same model, trained on different datasets, using $\mathcal{L_{\text{mark}}}$ between the markpainted patch and the target. Results are averaged across all possible input-mask-target combinations. Notice the large $\pm\sigma$ error bars, indicating high variability in transferability depending on the input combination.}
    \label{fig:dataset_transferability}
\end{figure}

\subsection{Transferability of Targeted Markpainted Examples}
\label{sec:eval:transfer}

Here we investigate the transferability of markpainted images in a blind black-box scenario, using previously-constructed markpainted samples to fool other models without knowledge of their internals. We investigate transferability across both model architectures and datasets. We report mean model performance with an increasing perturbation budget. Although an increased budget helps with transferability, the improvement is marginal in most cases. Note that variances of the measurements here are large; they reflect differences in input images and how different color targets transfer across models. 

In \Cref{fig:transferability}, we turn to the question of whether inpainter models show greater transferability if pretrained with the same dataset. RN and CRFILL, both trained on Places2, demonstrate a correlated decreasing pattern on the right plot of \Cref{fig:transferability}, showing that inpainters trained on the same dataset might suffer from transferred markpainting samples.

In~\Cref{fig:dataset_transferability} we demonstrate the transferability of markpainted examples within the same model architecture trained on different datasets. For this experiment we use EdgeConnect -- trained on CelebA, Paris StreetView, and Places2 -- and RFR -- trained on CelebA and Paris StreetView. Effectiveness of markpainted examples degrades to varying degrees when used by a model trained on a different dataset. However, the graphs demonstrate that markpainted examples are transferable within the same model architecture.

In general, we found that transferability exists across different inpainter models and datasets. This is broadly equivalent to the robustness of a watermark protected by our technique. It shows greater transferability when inpainters are trained on the same dataset or share the model architecture.

\begin{figure}[t]
    \centering
    \begin{subfigure}[t]{0.49\linewidth}
      \centering
      \includegraphics[width=\linewidth]{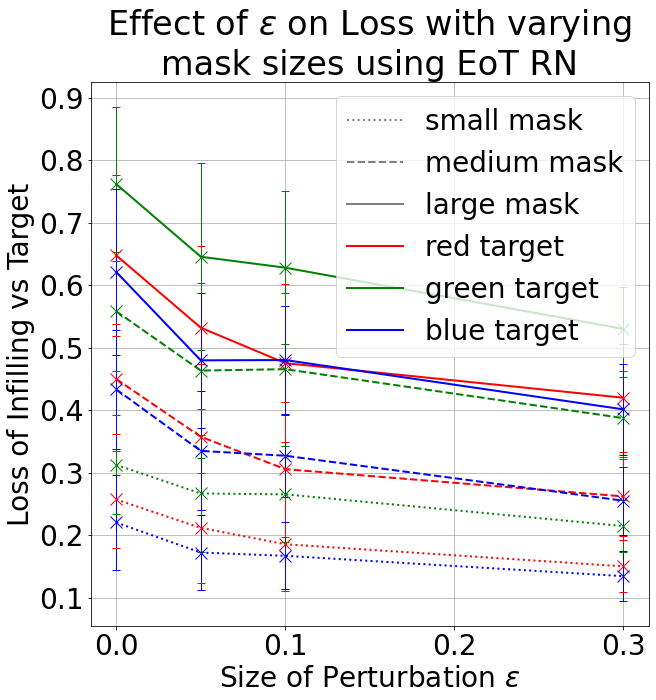}
      \caption{\scriptsize{EoT technique on RN.}}
    \end{subfigure}
    \begin{subfigure}[t]{0.49\linewidth}
      \centering
      \includegraphics[width=\linewidth]{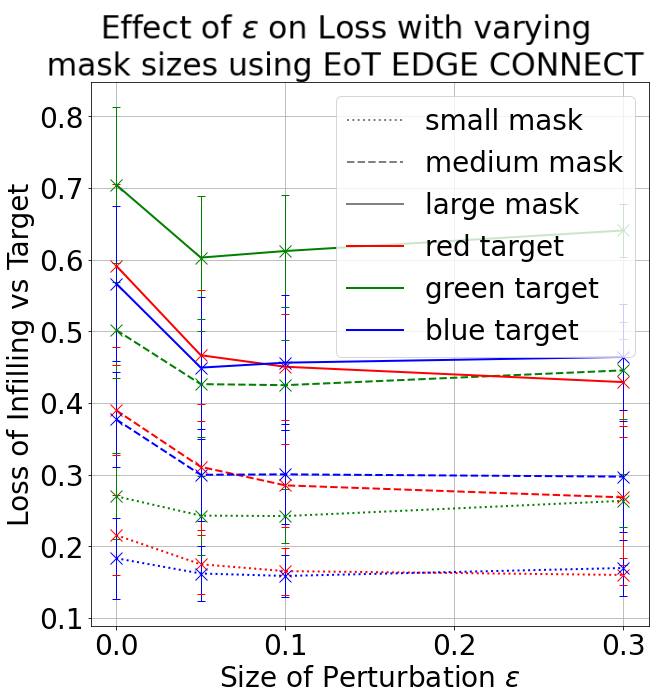}
      \caption{\scriptsize{EoT technique on EdgeConnect.}}
    \end{subfigure}
    \caption{Effect of $\epsilon$ on the effectiveness of the proposed mask-agnostic markpainting technique. Results are averaged across the \emph{places\_subset16} images, with $\pm\sigma$ error bars. It is evident that the technique's effectiveness is very much architecture dependent, with a high sensitivity to the input.}
    \label{fig:eot}
\end{figure}

\begin{figure*}[t]
    \centering
    \begin{subfigure}[t]{0.49\linewidth}
      \centering
      \includegraphics[width=\linewidth]{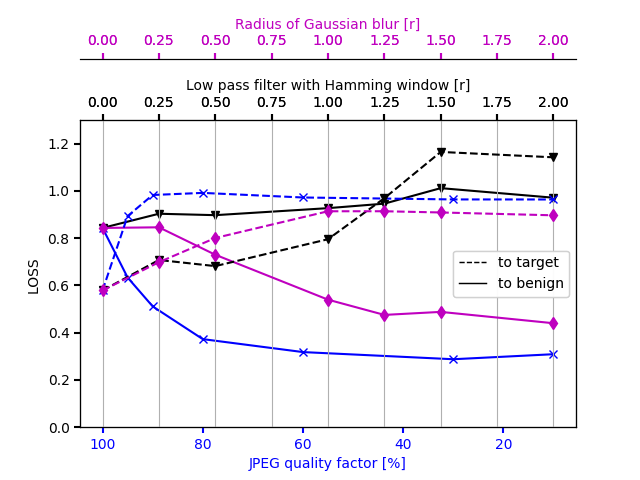}
      \caption{JPEG compression, low pass filtering and Gaussian blur. }
    \end{subfigure}
    \begin{subfigure}[t]{0.49\linewidth}
      \centering
      \includegraphics[width=\linewidth]{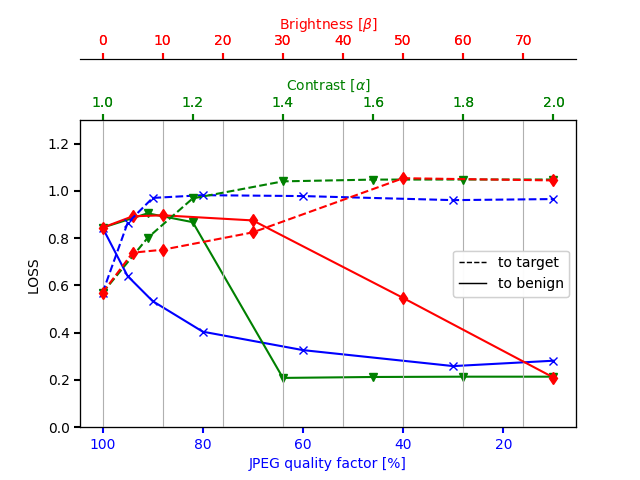}
      \caption{JPEG compression, brightness and contrast. }
    \end{subfigure}
    \caption{Effect of different transformation-based defenses. Results are from performing these transformations on the watermark example presented in~\Cref{fig:watermark_crfill}.}
    \label{fig:defence}
\end{figure*}

\subsection{Mask-agnostic Markpainting}
\label{sec:eval:agnostic}

\Cref{fig:eot} shows the effectiveness of our EoT method for mask-agnostic markpainting. The number of iterations was taken to be $1500$ with a step size of $\frac{\epsilon}{30}$, with $m_\text{min} = 0.01$ and $m_\text{max} = 0.1$. The evaluation masks are taken to be fixed random masks covering 2.5\%, 5\% and 10\% of the image.
The effectiveness was found to be architecture dependent. Moreover, certain images were more susceptible to markpainting than others. Investigating the exact causes of this architecture and image dependence is left to future work.


\section{Discussion}

\subsection{Countering Markpainting}

We have shown that modern inpainters can be manipulated to inpaint arbitrary target images. This naturally leads to the question of how one can counter markpainting. As markpainting aims to be explicitly imperceptible, it usually does not disrupt lower parts of the frequency spectra responsible for sharp edges, instead concentrating on the higher-frequency components. We propose a mechanism which accounts for this.

We find that transformation-based manipulations work relatively well in countering markpainting. \Cref{fig:defence} shows dissimilarity $\mathcal{L_{\text{mark}}}$ between the markpainted patch and the target/benign images. We test five different transformations: JPEG compression, low-pass filtering, Gaussian blurring, contrast adjustments, and brightness adjustments. Each manipulation significantly reduced markpainting performance, but had different impact on the inpainting performance in the benign cases. Simple low-pass filtering reduces similarity of the reconstruction to the target image, but also causes significant deviation from the benign reconstruction. This highlights the trade-off between countering markpainting and preserving the benign inpainted patch. Although some transformations decrease the performance of markpainting, they change the original image significantly as well. Thus, manual human involvement appears to be required, which is highly likely to limit the scalability of abuse based on inpainting.  

\subsection{Interpretability of Markpainting}

Unlike adversarial examples for classification tasks, markpainting can be interpreted. Indeed, we find that we could often visually tell what an increased perturbation budget was changing in our perception of the inpainter model. Evaluation suggests that although markpainting can be made transferable, it usually is not. We find that it is, perhaps intuitively, harder to markpaint colors or shapes that are not present in the original image. Complex shapes, and contours that do not naturally extend from the mask's boundaries, also prove to be a challenge. In contrast, models that have been trained on the CelebA dataset are easier to fool into markpainting faces, as demonstrated strikingly in a visualization provided in Section A of our Appendix.




\section{Conclusion}

We introduce the idea of \emph{markpainting}: fooling an inpainting system into generating a patch similar to an arbitrary target. Moreover, we demonstrate through mask-agnostic markpainting that the technique does not need to be restricted to a particular mask to be effective. We also show the existence of some degree of transferability of these adversarial examples both within a single model and between different model architectures. 

Markpainting has wide implications. Image owners can now protect their digital assets with less removable watermarks, or treat them so that any later manipulation such as object removal becomes easier to detect.


\section*{Acknowledgments}

We thank the reviewers for their insightful feedback. We want to explicitly thank Mohammad Yaghini, Stephan Rabanser, Gabriel Deza, Natalie Dullerud, Ali Shahin Shamsabadi and Nicholas Boucher for their help and comments. This work was supported by CIFAR (through a Canada CIFAR AI Chair), by EPSRC, by Apple, by Bosch Forschungsstiftung im Stifterverband, by NSERC, and by a gift from Microsoft. We also thank the Vector Institute's sponsors.

\bibliography{bibliography}
\bibliographystyle{icml2021}

\appendix

\section{Applying Markpainting to a Collection of Models}

\begin{figure}[!h]
    \centering
    \includegraphics[width=0.85\linewidth]{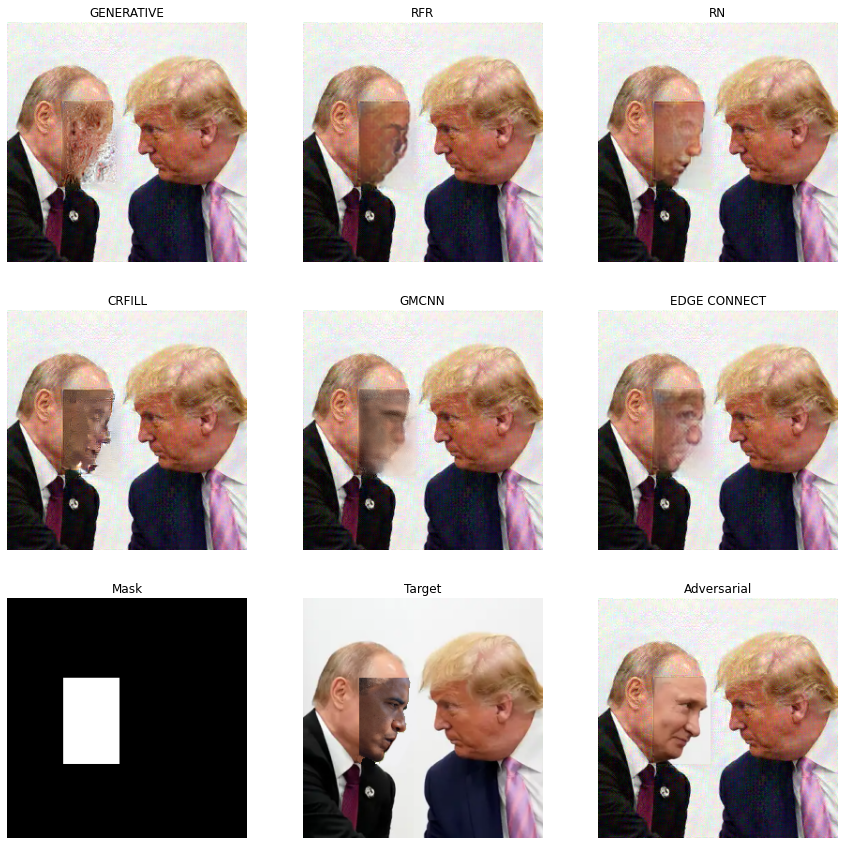}
    \caption{Markpainting Vladimir Putin's face with former President Obama's. The mix of colors in the image makes it significantly easier to attack. Attack performed over $500$ iterations, with $\epsilon=0.1$. The original photo with Putin and Trump is taken from \href{https://www.theguardian.com/world/2020/jun/30/trump-putin-russia-afghanistan-us-soliders-bounty}{The Guardian}. Photograph of Obama taken from \href{https://www.acclaimimages.com/_gallery/_pages/0519-1010-0716-5355.html}{Acclaim Images}.}
    \label{fig:example_putin_trump}
\end{figure}

\begin{figure}[!h]
    \centering
    \includegraphics[width=0.85\linewidth]{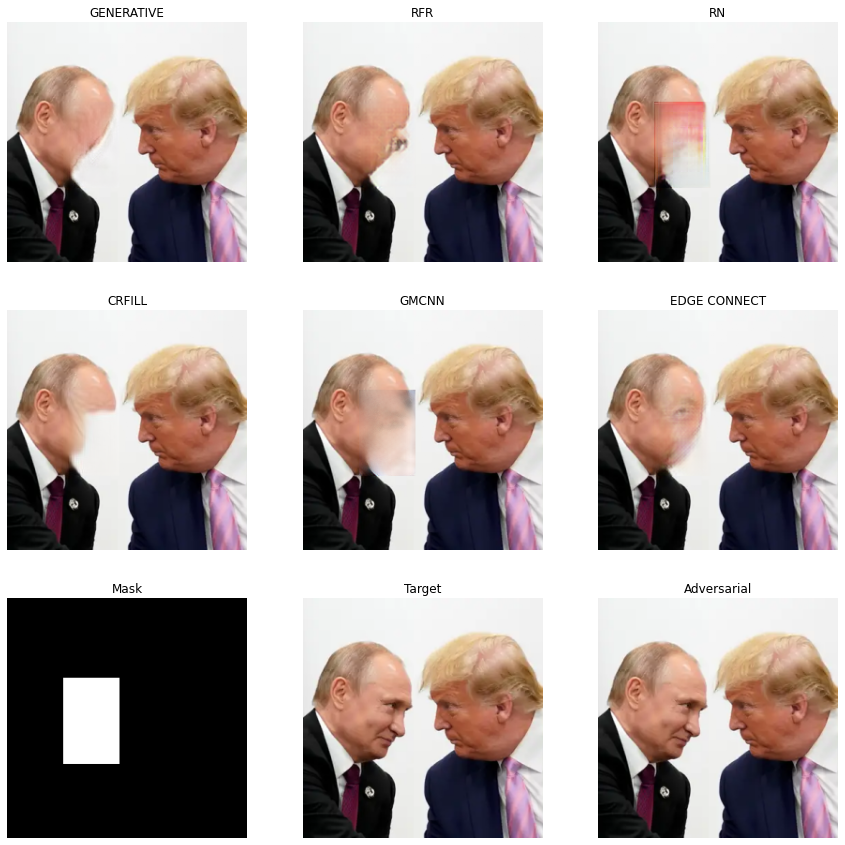}
    \caption{Benign infilling of the example image in~\Cref{fig:example_putin_trump}.}
    \label{fig:example_putin_trump_benign}
\end{figure}

As illustrated in the markpainting algorithm, 
we find that we are able to markpaint a collection of models $\mathbf{\Theta}$ simultaneously using a single input image $\mathbf{I}$ as detailed in Algorithm 1 in the paper. 
This means a single adversarial image serves as an input to all considered models.
An example application of this in a white-box setting is shown in \Cref{fig:example_putin_trump}
, where the same sample is perturbed to be producing a different face after inpainting by a 6 different inpainters. 
\Cref{fig:example_putin_trump_benign} shows the benign samples from the inpainters, these are the infilling effect if there are no markpainting generated noises.
The inpainter will naively fill the masked region without facial details, but the generated markpainted example can influence this filling to provide facial details of Obama.


\Cref{tab:all_model} shows the details of how this technique works with different inpainters. In \Cref{tab:all_model}, the adversarial samples are generated using all models and evaluated on each model individually.
The table reports the loss, L2 norms, peak signal to noise ratio (PSNR) and structural similarity index measure (SSIM) for accessing the inpainted image quality.


\section{Targeted Application of Inpainting}
\Cref{tab:individual_model} shows the results of the markpainting technique on different inpainter models and it is an extended version of Table 2 in the main paper.
The markpainting technique is launched at each individual inpainter and evaluated on that inpainter with different perturbation budgets.

The table reports the loss, L2 norms, peak signal to noise ratio (PSNR) and structural index similarity (SSIM) for accessing the inpainted image quality.


\section{Transferability of Markpainting}
In \Cref{tab:single_model}, we assess the transferability of markpainting.

\begin{figure*}[t]
    \centering
    \includegraphics[width=\textwidth]{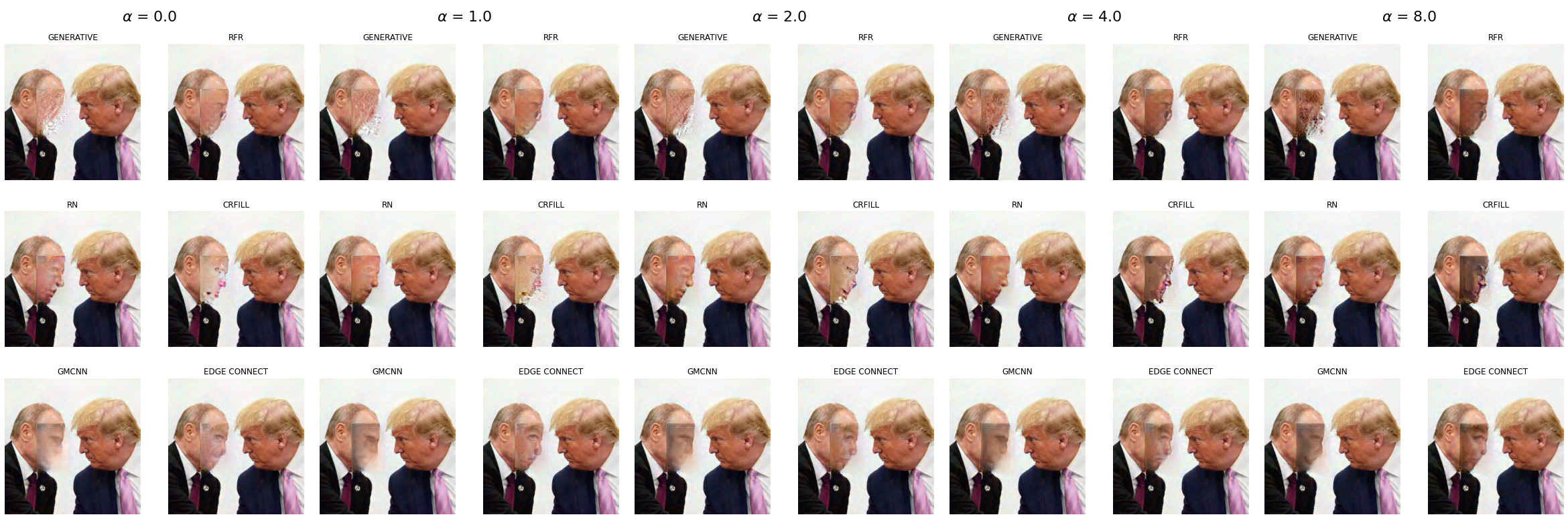}
    \caption{Effect of $\alpha$ on markpainting. $\alpha \in \{0, 1, 2, 4, 8\}$.}
    \label{fig:alpha_effect}
\end{figure*}

\section{Parameter Choices}
In the paper, we provided a visualization of having an increasing $\epsilon$ budgets in Figure 4. This term controls a loss trade-off between the network loss and the L2 loss.
As we can see, in \Cref{fig:alpha_effect}, when the $\alpha$ value increases, the markpainted image gets closer to the target.
We also show the original benign inpainting results in \Cref{fig:example_putin_trump_benign} as a baseline for comparisons. 
It is worth to mention that the baseline simply fills the face with surrounding colors.

We also further study the effect of $\epsilon$ of the markpainting technique. In the evaluation, the effect of different epsilons are shown for the RN inpainter, we further illustrate the effect of epsilons on other inpainters (RFR and CRFILL), and they are \Cref{fig:eps_painter_crfill} and \Cref{fig:eps_painter_rfr} respectively.

\section{Evaluation Details}

The \emph{places\_subset16} dataset that we used to evaluate our proposed method on -- a series of 16 randomly-selected images from the Places2 dataset~\cite{zhou2017places} -- is visualized in~\Cref{fig:places_16}.

\begin{figure}[!h]
	\centering
	\begin{subfigure}[b]{\linewidth}
        \centering
        \includegraphics[width=\linewidth]{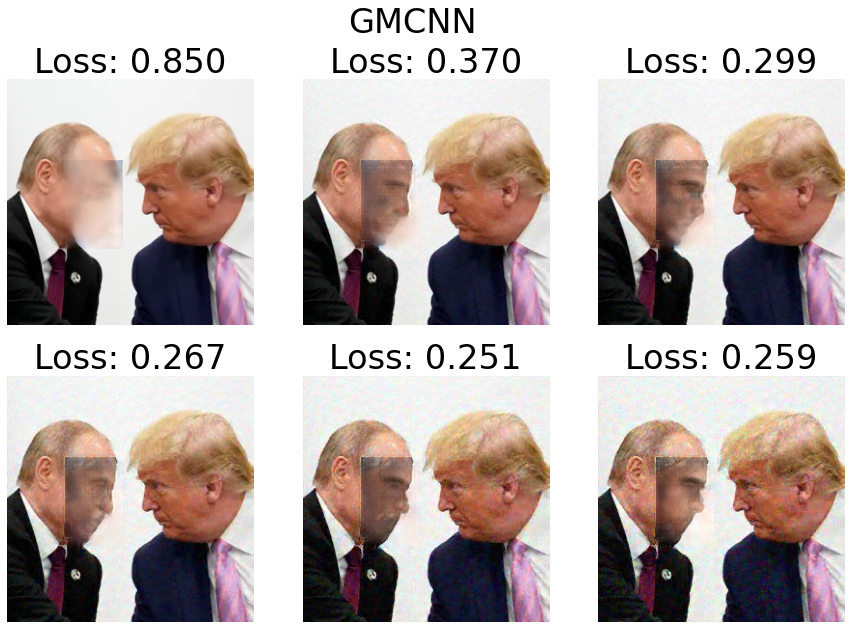}
        \caption{Applied to the GMCNN inpainter model.}
    \end{subfigure}
    \begin{subfigure}[b]{\linewidth}
        \centering
        \includegraphics[width=\linewidth]{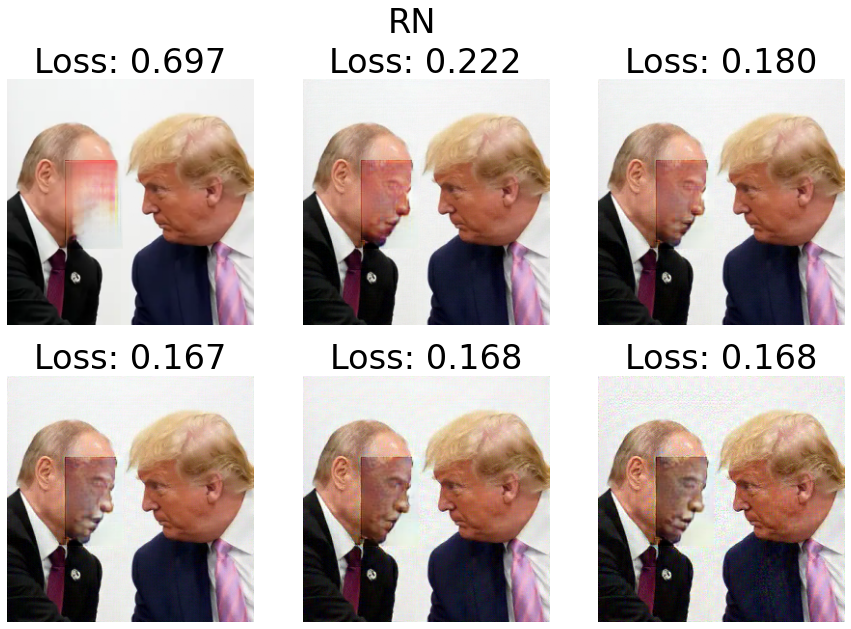}
        \caption{Applied to the RN inpainter model.}
    \end{subfigure}
	\caption{Correspondence between numeric loss to target image and the obtained markpainted results for a complex target.}\label{fig:corr_putin}
\end{figure}

\begin{figure}[h]
	\centering
	\begin{subfigure}[b]{\linewidth}
        \centering
        \includegraphics[width=\linewidth]{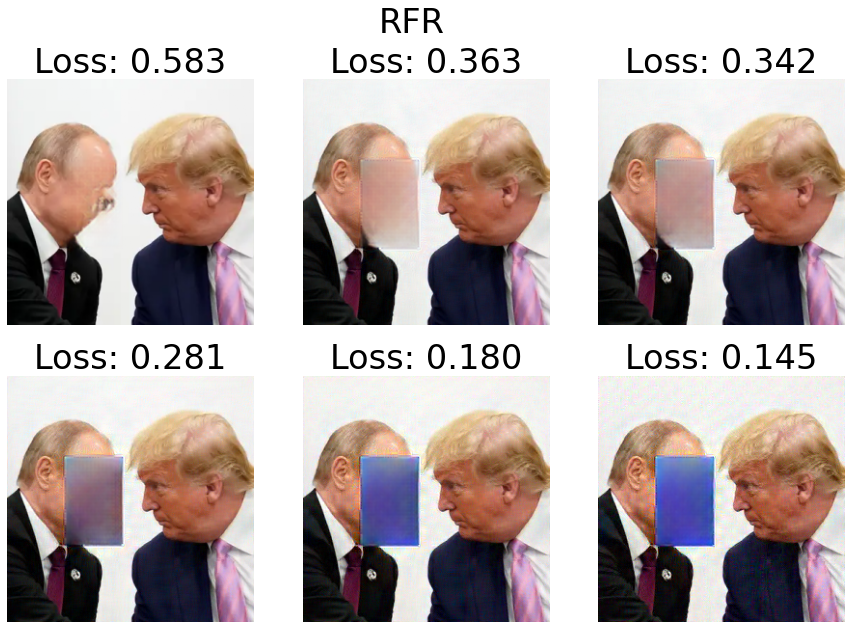}
        \caption{Applied to the RFR inpainter model.}
    \end{subfigure}
    \begin{subfigure}[b]{\linewidth}
        \centering
        \includegraphics[width=\linewidth]{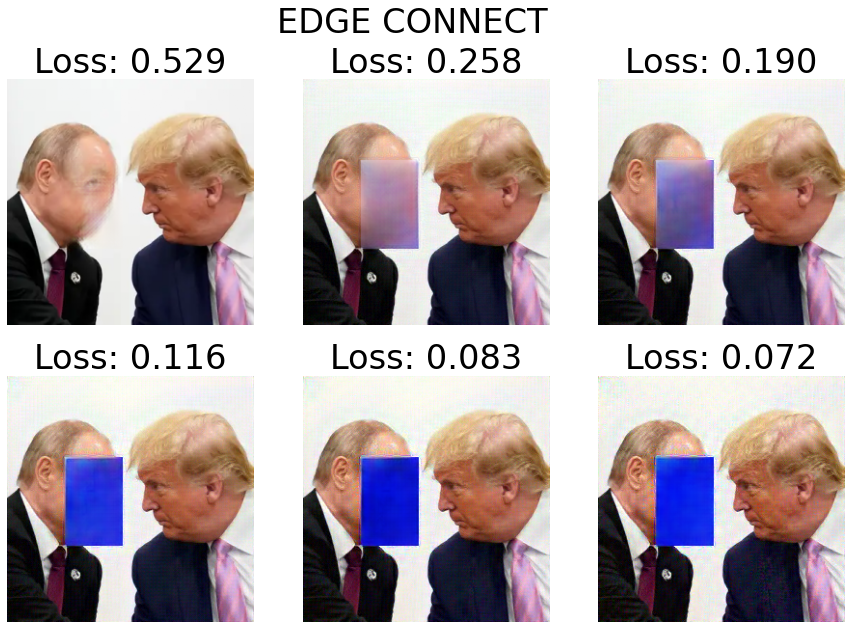}
        \caption{Applied to the EDGE CONNECT inpainter model.}
    \end{subfigure}
	\caption{Correspondence between numeric loss to target image and the obtained markpainted results for a solid-color target.}\label{fig:corr_colour}
\end{figure}

We understand that it is of interest to readers to be able to visualize the numeric loss values we quote in our results. In~\Cref{fig:corr_putin}, we present visual examples of how numeric loss relates to the markpainted results for complex targets; and in~\Cref{fig:corr_colour} we do the same for a solid-color target.



\newpage
\section{Watermark Removal}

\begin{figure}[t]
    \centering
    \includegraphics[width=0.9\linewidth]{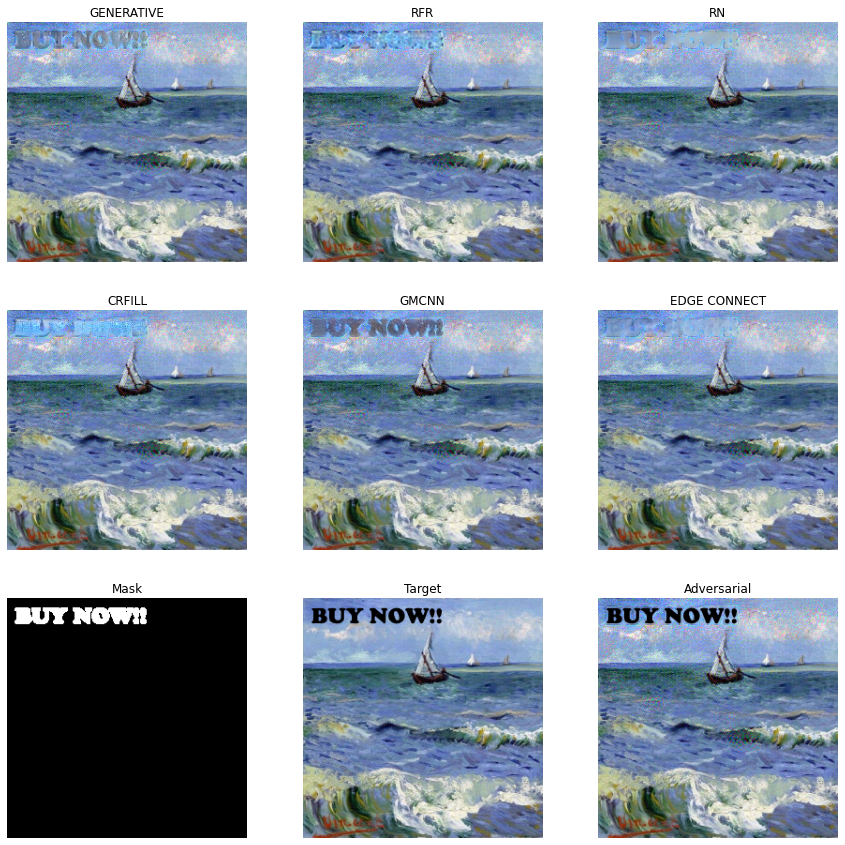}
    \caption{Example usage of the methods proposed in this paper: watermark encoding into an image using perceptually indistinguishable noise ($\epsilon=0.15$). In this example, GENERATIVE is markpainted, but the intentionally disruptive result is transferred to all models. Image from~\cite{yu2018generative}.}\label{fig:watermark}
\end{figure}

\begin{figure*}[t]
	\centering
	\includegraphics[width=\textwidth]{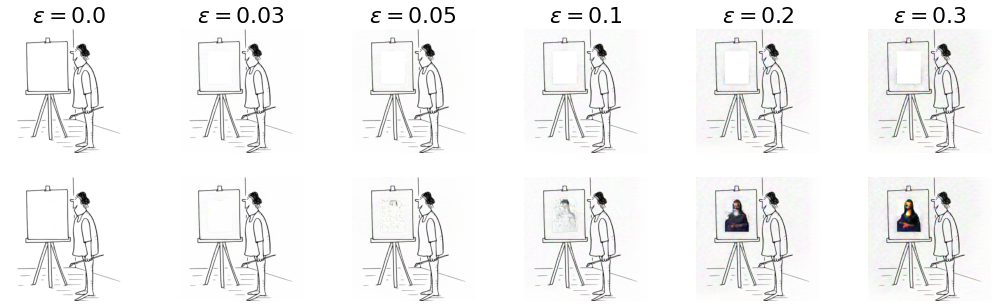}
	\caption{Inpainting with increasing perturbation epsilon budget. Top row is the adversarial images generated using markpainting, and second row is the inpainted results of these adversarial images. We target the CRFILL inpainter with 500 iterations and a step size of $\epsilon/100$. Note that this example is really hard, because we are filling a black and white image with color.}
	\label{fig:eps_painter_crfill}
\end{figure*}

\begin{figure*}[t]
	\centering
	\includegraphics[width=\textwidth]{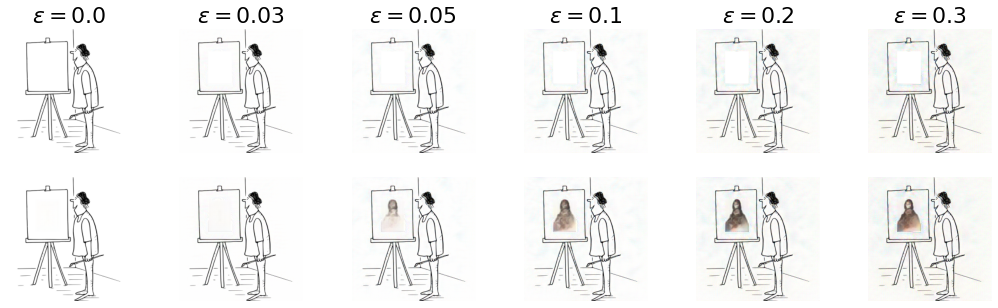}
	\caption{Inpainting with increasing perturbation epsilon budget. Top row is the adversarial images generated using markpainting, and second row is the inpainted results of these adversarial images. We target the RFR inpainter with 500 iterations and a step size of $\epsilon/100$. Note that this example is really hard, because we are filling a black and white image with color.}
	\label{fig:eps_painter_rfr}
\end{figure*}

\begin{figure*}[h]
    \centering
    \begin{subfigure}[b]{0.25\linewidth}
        \centering
        \includegraphics[width=\linewidth]{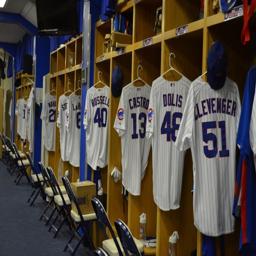}
    \end{subfigure}%
    \begin{subfigure}[b]{0.25\linewidth}
        \centering
        \includegraphics[width=\linewidth]{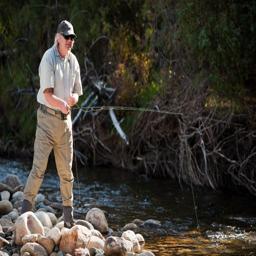}
    \end{subfigure}%
    \begin{subfigure}[b]{0.25\linewidth}
        \centering
        \includegraphics[width=\linewidth]{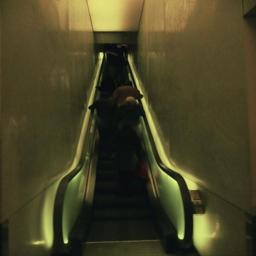}
    \end{subfigure}%
    \begin{subfigure}[b]{0.25\linewidth}
        \centering
        \includegraphics[width=\linewidth]{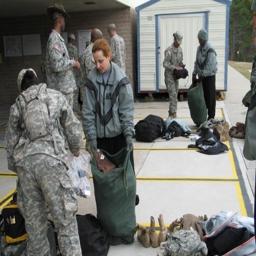}
    \end{subfigure}
    \begin{subfigure}[b]{0.25\linewidth}
        \centering
        \includegraphics[width=\linewidth]{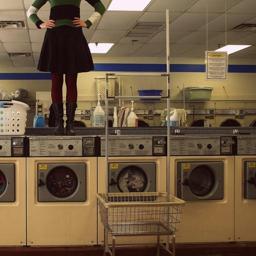}
    \end{subfigure}%
    \begin{subfigure}[b]{0.25\linewidth}
        \centering
        \includegraphics[width=\linewidth]{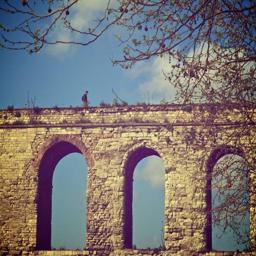}
    \end{subfigure}%
    \begin{subfigure}[b]{0.25\linewidth}
        \centering
        \includegraphics[width=\linewidth]{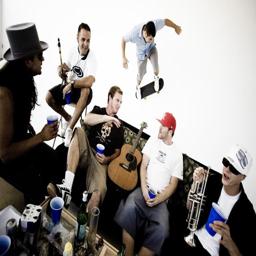}
    \end{subfigure}%
    \begin{subfigure}[b]{0.25\linewidth}
        \centering
        \includegraphics[width=\linewidth]{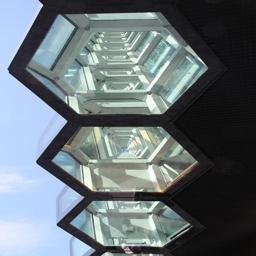}
    \end{subfigure}
    \begin{subfigure}[b]{0.25\linewidth}
        \centering
        \includegraphics[width=\linewidth]{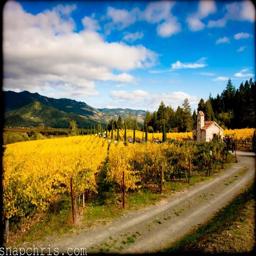}
    \end{subfigure}%
    \begin{subfigure}[b]{0.25\linewidth}
        \centering
        \includegraphics[width=\linewidth]{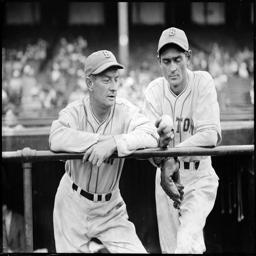}
    \end{subfigure}%
    \begin{subfigure}[b]{0.25\linewidth}
        \centering
        \includegraphics[width=\linewidth]{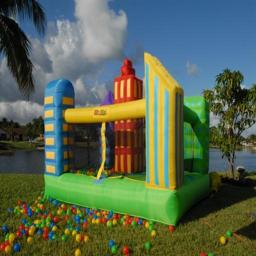}
    \end{subfigure}%
    \begin{subfigure}[b]{0.25\linewidth}
        \centering
        \includegraphics[width=\linewidth]{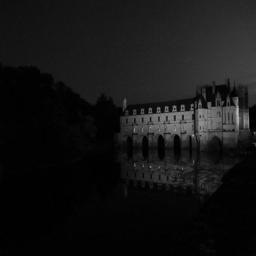}
    \end{subfigure}
    \begin{subfigure}[b]{0.25\linewidth}
        \centering
        \includegraphics[width=\linewidth]{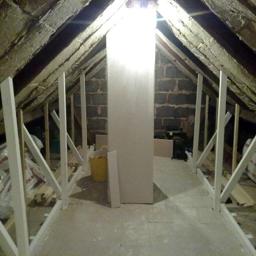}
    \end{subfigure}%
    \begin{subfigure}[b]{0.25\linewidth}
        \centering
        \includegraphics[width=\linewidth]{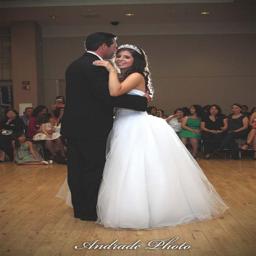}
    \end{subfigure}%
    \begin{subfigure}[b]{0.25\linewidth}
        \centering
        \includegraphics[width=\linewidth]{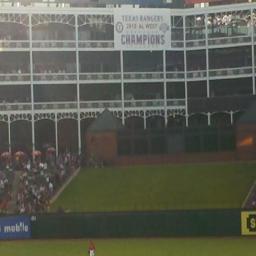}
    \end{subfigure}%
    \begin{subfigure}[b]{0.25\linewidth}
        \centering
        \includegraphics[width=\linewidth]{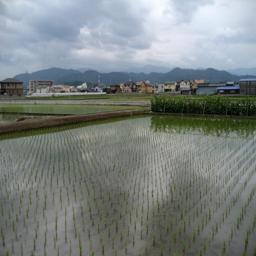}
    \end{subfigure}%
    \caption{The \emph{places\_subset16} dataset used for evaluation. A subset of Places2~\cite{zhou2017places}.}\label{fig:places_16}
\end{figure*}

We show more results on the watermark removal with $\epsilon=0.15$. 
The objective is to build an image that is resistant to watermark removals using markpainting.
\Cref{fig:watermark} shows that markpainted images (top two rows) are in general more robust to different inpainters trying to fill the watermark.



\begin{table*}[t]
\begin{adjustbox}{scale=0.5,center}
\begin{tabular}{@{}llrrrr|rrrr|rrrr@{}}
\toprule
\multicolumn{2}{l}{Distance to:} &\multicolumn{4}{c|}{Original} & \multicolumn{4}{c|}{Adversarial Target} & \multicolumn{4}{c}{Benign} \\
$\epsilon$ & &Loss & $l_2$ & PSNR & SSIM & Loss & $l_2$ & PSNR & SSIM & Loss & $l_2$ & PSNR & SSIM\\
\midrule
\multirow{5}{*}{$ 0.0 $}
&GENERATIVE & 0.467 ($\pm$0.282) & 0.111 ($\pm$0.113) & 12.491 ($\pm$6.257) & 0.250 ($\pm$0.222) & 0.755 ($\pm$0.334) & 1.186 ($\pm$0.298) & -0.622 ($\pm$0.989) & 0.036 ($\pm$0.100) & 0.179 ($\pm$0.062) & 0.000 ($\pm$0.000) & 134.177 ($\pm$3.400) & 1.000 ($\pm$0.000)\\
&RFR & 0.279 ($\pm$0.220) & 0.027 ($\pm$0.034) & 19.156 ($\pm$7.224) & 0.387 ($\pm$0.261) & 0.433 ($\pm$0.191) & 0.292 ($\pm$0.072) & 5.464 ($\pm$0.992) & 0.090 ($\pm$0.055) & 0.001 ($\pm$0.001) & 0.000 ($\pm$0.000) & 144.860 ($\pm$4.024) & 1.000 ($\pm$0.000)\\
&RN & 0.300 ($\pm$0.227) & 0.025 ($\pm$0.023) & 17.891 ($\pm$4.161) & 0.420 ($\pm$0.220) & 0.473 ($\pm$0.204) & 0.292 ($\pm$0.060) & 5.438 ($\pm$0.851) & 0.104 ($\pm$0.051) & 0.001 ($\pm$0.001) & 0.000 ($\pm$0.000) & inf ($\pm$nan) & 1.000 ($\pm$0.000)\\
&CRFILL & 0.470 ($\pm$0.295) & 0.109 ($\pm$0.120) & 13.342 ($\pm$7.767) & 0.319 ($\pm$0.257) & 0.796 ($\pm$0.355) & 1.205 ($\pm$0.309) & -0.683 ($\pm$1.021) & 0.044 ($\pm$0.115) & 0.180 ($\pm$0.059) & 0.000 ($\pm$0.000) & inf ($\pm$nan) & 1.000 ($\pm$0.000)\\
&GMCNN & 0.485 ($\pm$0.267) & 0.122 ($\pm$0.108) & 10.891 ($\pm$4.560) & 0.210 ($\pm$0.208) & 0.721 ($\pm$0.289) & 1.136 ($\pm$0.283) & -0.432 ($\pm$1.014) & 0.047 ($\pm$0.110) & 0.181 ($\pm$0.060) & 0.000 ($\pm$0.000) & inf ($\pm$nan) & 1.000 ($\pm$0.000)\\
&EDGE CONNECT & 0.290 ($\pm$0.226) & 0.025 ($\pm$0.022) & 18.198 ($\pm$5.129) & 0.390 ($\pm$0.245) & 0.422 ($\pm$0.189) & 0.299 ($\pm$0.085) & 5.383 ($\pm$1.088) & 0.104 ($\pm$0.056) & 0.001 ($\pm$0.001) & 0.000 ($\pm$0.000) & 135.065 ($\pm$3.821) & 1.000 ($\pm$0.000)\\
\midrule
\multirow{5}{*}{$ 0.03 $}
&GENERATIVE & 0.435 ($\pm$0.249) & 0.104 ($\pm$0.110) & 12.627 ($\pm$5.859) & 0.247 ($\pm$0.212) & 0.663 ($\pm$0.268) & 1.147 ($\pm$0.282) & -0.481 ($\pm$0.966) & 0.040 ($\pm$0.098) & 0.254 ($\pm$0.109) & 0.013 ($\pm$0.011) & 21.106 ($\pm$5.096) & 0.638 ($\pm$0.117)\\
&RFR & 0.298 ($\pm$0.209) & 0.031 ($\pm$0.038) & 17.069 ($\pm$4.168) & 0.346 ($\pm$0.242) & 0.357 ($\pm$0.147) & 0.259 ($\pm$0.069) & 6.003 ($\pm$1.086) & 0.105 ($\pm$0.058) & 0.102 ($\pm$0.076) & 0.008 ($\pm$0.006) & 22.280 ($\pm$3.564) & 0.775 ($\pm$0.122)\\
&RN & 0.328 ($\pm$0.207) & 0.042 ($\pm$0.027) & 14.641 ($\pm$2.940) & 0.343 ($\pm$0.192) & 0.299 ($\pm$0.142) & 0.221 ($\pm$0.058) & 6.724 ($\pm$1.233) & 0.171 ($\pm$0.066) & 0.183 ($\pm$0.089) & 0.023 ($\pm$0.017) & 17.508 ($\pm$3.191) & 0.604 ($\pm$0.127)\\
&CRFILL & 0.446 ($\pm$0.251) & 0.124 ($\pm$0.149) & 11.034 ($\pm$4.111) & 0.232 ($\pm$0.193) & 0.642 ($\pm$0.279) & 1.065 ($\pm$0.289) & -0.134 ($\pm$1.081) & 0.062 ($\pm$0.105) & 0.315 ($\pm$0.126) & 0.044 ($\pm$0.044) & 14.935 ($\pm$3.315) & 0.497 ($\pm$0.139)\\
&GMCNN & 0.549 ($\pm$0.286) & 0.554 ($\pm$0.704) & 5.127 ($\pm$4.498) & 0.038 ($\pm$0.236) & 0.611 ($\pm$0.275) & 1.200 ($\pm$0.656) & -0.301 ($\pm$1.938) & 0.046 ($\pm$0.188) & 0.358 ($\pm$0.187) & 0.439 ($\pm$0.681) & 7.602 ($\pm$5.764) & 0.182 ($\pm$0.323)\\
&EDGE CONNECT & 0.305 ($\pm$0.209) & 0.036 ($\pm$0.020) & 15.090 ($\pm$2.432) & 0.343 ($\pm$0.212) & 0.305 ($\pm$0.135) & 0.252 ($\pm$0.080) & 6.171 ($\pm$1.201) & 0.157 ($\pm$0.058) & 0.105 ($\pm$0.067) & 0.017 ($\pm$0.012) & 18.685 ($\pm$3.159) & 0.715 ($\pm$0.126)\\
\midrule
\multirow{5}{*}{$ 0.05 $}
&GENERATIVE & 0.436 ($\pm$0.248) & 0.103 ($\pm$0.109) & 12.499 ($\pm$5.582) & 0.237 ($\pm$0.209) & 0.639 ($\pm$0.254) & 1.119 ($\pm$0.273) & -0.375 ($\pm$0.961) & 0.044 ($\pm$0.097) & 0.277 ($\pm$0.125) & 0.022 ($\pm$0.020) & 18.740 ($\pm$4.933) & 0.512 ($\pm$0.150)\\
&RFR & 0.323 ($\pm$0.218) & 0.037 ($\pm$0.038) & 15.711 ($\pm$3.484) & 0.323 ($\pm$0.228) & 0.323 ($\pm$0.133) & 0.230 ($\pm$0.067) & 6.542 ($\pm$1.202) & 0.125 ($\pm$0.060) & 0.147 ($\pm$0.098) & 0.017 ($\pm$0.013) & 18.790 ($\pm$3.099) & 0.666 ($\pm$0.137)\\
&RN & 0.357 ($\pm$0.218) & 0.052 ($\pm$0.033) & 13.757 ($\pm$2.969) & 0.323 ($\pm$0.195) & 0.223 ($\pm$0.110) & 0.177 ($\pm$0.056) & 7.734 ($\pm$1.431) & 0.203 ($\pm$0.060) & 0.239 ($\pm$0.118) & 0.036 ($\pm$0.022) & 15.286 ($\pm$2.971) & 0.514 ($\pm$0.155)\\
&CRFILL & 0.483 ($\pm$0.247) & 0.195 ($\pm$0.186) & 8.130 ($\pm$2.850) & 0.139 ($\pm$0.137) & 0.526 ($\pm$0.258) & 0.826 ($\pm$0.298) & 1.118 ($\pm$1.639) & 0.121 ($\pm$0.132) & 0.392 ($\pm$0.158) & 0.136 ($\pm$0.108) & 9.709 ($\pm$2.998) & 0.260 ($\pm$0.138)\\
&GMCNN & 0.579 ($\pm$0.315) & 0.687 ($\pm$0.764) & 4.007 ($\pm$4.579) & 0.020 ($\pm$0.225) & 0.601 ($\pm$0.281) & 1.281 ($\pm$0.769) & -0.418 ($\pm$2.293) & 0.040 ($\pm$0.224) & 0.399 ($\pm$0.205) & 0.567 ($\pm$0.729) & 5.714 ($\pm$5.406) & 0.111 ($\pm$0.306)\\
&EDGE CONNECT & 0.320 ($\pm$0.215) & 0.039 ($\pm$0.022) & 14.666 ($\pm$2.352) & 0.335 ($\pm$0.209) & 0.269 ($\pm$0.119) & 0.223 ($\pm$0.073) & 6.694 ($\pm$1.218) & 0.175 ($\pm$0.057) & 0.142 ($\pm$0.088) & 0.024 ($\pm$0.014) & 17.011 ($\pm$2.606) & 0.627 ($\pm$0.138)\\
\midrule
\multirow{5}{*}{$ 0.1 $}
&GENERATIVE & 0.445 ($\pm$0.249) & 0.107 ($\pm$0.110) & 11.825 ($\pm$4.632) & 0.210 ($\pm$0.198) & 0.595 ($\pm$0.235) & 1.036 ($\pm$0.256) & -0.037 ($\pm$0.976) & 0.063 ($\pm$0.098) & 0.309 ($\pm$0.144) & 0.040 ($\pm$0.031) & 15.478 ($\pm$3.927) & 0.349 ($\pm$0.169)\\
&RFR & 0.365 ($\pm$0.231) & 0.057 ($\pm$0.052) & 13.515 ($\pm$2.895) & 0.299 ($\pm$0.200) & 0.259 ($\pm$0.112) & 0.176 ($\pm$0.065) & 7.822 ($\pm$1.560) & 0.178 ($\pm$0.065) & 0.212 ($\pm$0.119) & 0.039 ($\pm$0.026) & 14.942 ($\pm$2.691) & 0.504 ($\pm$0.163)\\
&RN & 0.385 ($\pm$0.231) & 0.067 ($\pm$0.041) & 12.476 ($\pm$2.645) & 0.311 ($\pm$0.191) & 0.145 ($\pm$0.069) & 0.128 ($\pm$0.044) & 9.169 ($\pm$1.496) & 0.240 ($\pm$0.048) & 0.286 ($\pm$0.144) & 0.055 ($\pm$0.028) & 13.197 ($\pm$2.453) & 0.448 ($\pm$0.157)\\
&CRFILL & 0.546 ($\pm$0.268) & 0.384 ($\pm$0.222) & 4.727 ($\pm$2.177) & 0.064 ($\pm$0.111) & 0.319 ($\pm$0.172) & 0.403 ($\pm$0.189) & 4.404 ($\pm$2.039) & 0.333 ($\pm$0.183) & 0.486 ($\pm$0.218) & 0.358 ($\pm$0.180) & 4.965 ($\pm$2.107) & 0.089 ($\pm$0.113)\\
&GMCNN & 0.606 ($\pm$0.344) & 0.750 ($\pm$0.799) & 3.553 ($\pm$4.553) & 0.009 ($\pm$0.231) & 0.584 ($\pm$0.303) & 1.292 ($\pm$0.841) & -0.272 ($\pm$2.640) & 0.060 ($\pm$0.257) & 0.435 ($\pm$0.253) & 0.641 ($\pm$0.773) & 4.890 ($\pm$5.185) & 0.065 ($\pm$0.289)\\
&EDGE CONNECT & 0.350 ($\pm$0.228) & 0.049 ($\pm$0.027) & 13.838 ($\pm$2.581) & 0.320 ($\pm$0.205) & 0.212 ($\pm$0.093) & 0.181 ($\pm$0.065) & 7.659 ($\pm$1.361) & 0.201 ($\pm$0.051) & 0.193 ($\pm$0.115) & 0.036 ($\pm$0.021) & 15.168 ($\pm$2.732) & 0.529 ($\pm$0.150)\\
\midrule
\multirow{5}{*}{$ 0.2 $}
&GENERATIVE & 0.468 ($\pm$0.259) & 0.134 ($\pm$0.123) & 10.145 ($\pm$3.532) & 0.158 ($\pm$0.171) & 0.513 ($\pm$0.209) & 0.860 ($\pm$0.223) & 0.783 ($\pm$1.036) & 0.104 ($\pm$0.104) & 0.352 ($\pm$0.170) & 0.081 ($\pm$0.059) & 11.906 ($\pm$2.875) & 0.208 ($\pm$0.157)\\
&RFR & 0.411 ($\pm$0.243) & 0.087 ($\pm$0.053) & 11.230 ($\pm$2.331) & 0.267 ($\pm$0.167) & 0.187 ($\pm$0.084) & 0.117 ($\pm$0.052) & 9.788 ($\pm$2.117) & 0.229 ($\pm$0.060) & 0.272 ($\pm$0.139) & 0.073 ($\pm$0.037) & 11.902 ($\pm$2.202) & 0.382 ($\pm$0.151)\\
&RN & 0.407 ($\pm$0.244) & 0.092 ($\pm$0.054) & 11.054 ($\pm$2.421) & 0.291 ($\pm$0.187) & 0.101 ($\pm$0.042) & 0.091 ($\pm$0.031) & 10.679 ($\pm$1.520) & 0.264 ($\pm$0.045) & 0.314 ($\pm$0.159) & 0.079 ($\pm$0.037) & 11.511 ($\pm$2.128) & 0.404 ($\pm$0.153)\\
&CRFILL & 0.580 ($\pm$0.295) & 0.495 ($\pm$0.225) & 3.458 ($\pm$1.875) & 0.045 ($\pm$0.105) & 0.237 ($\pm$0.132) & 0.241 ($\pm$0.104) & 6.624 ($\pm$2.065) & 0.566 ($\pm$0.162) & 0.526 ($\pm$0.252) & 0.483 ($\pm$0.199) & 3.528 ($\pm$1.795) & 0.061 ($\pm$0.108)\\
&GMCNN & 0.615 ($\pm$0.373) & 0.768 ($\pm$0.803) & 3.436 ($\pm$4.543) & 0.013 ($\pm$0.227) & 0.547 ($\pm$0.319) & 1.239 ($\pm$0.912) & 0.178 ($\pm$3.046) & 0.093 ($\pm$0.296) & 0.446 ($\pm$0.275) & 0.659 ($\pm$0.765) & 4.698 ($\pm$5.162) & 0.044 ($\pm$0.265)\\
&EDGE CONNECT & 0.388 ($\pm$0.243) & 0.071 ($\pm$0.039) & 12.092 ($\pm$2.348) & 0.291 ($\pm$0.184) & 0.163 ($\pm$0.068) & 0.136 ($\pm$0.056) & 8.964 ($\pm$1.622) & 0.212 ($\pm$0.047) & 0.241 ($\pm$0.137) & 0.061 ($\pm$0.033) & 12.766 ($\pm$2.416) & 0.433 ($\pm$0.144)\\
\midrule
\multirow{5}{*}{$ 0.3 $}
&GENERATIVE & 0.494 ($\pm$0.269) & 0.174 ($\pm$0.131) & 8.528 ($\pm$2.806) & 0.112 ($\pm$0.143) & 0.445 ($\pm$0.197) & 0.709 ($\pm$0.205) & 1.654 ($\pm$1.164) & 0.146 ($\pm$0.121) & 0.387 ($\pm$0.190) & 0.127 ($\pm$0.075) & 9.608 ($\pm$2.286) & 0.140 ($\pm$0.132)\\
&RFR & 0.439 ($\pm$0.249) & 0.114 ($\pm$0.057) & 9.917 ($\pm$2.066) & 0.243 ($\pm$0.155) & 0.154 ($\pm$0.071) & 0.089 ($\pm$0.045) & 11.242 ($\pm$2.689) & 0.257 ($\pm$0.064) & 0.304 ($\pm$0.147) & 0.102 ($\pm$0.046) & 10.379 ($\pm$1.997) & 0.329 ($\pm$0.141)\\
&RN & 0.418 ($\pm$0.247) & 0.106 ($\pm$0.058) & 10.338 ($\pm$2.263) & 0.280 ($\pm$0.183) & 0.087 ($\pm$0.039) & 0.076 ($\pm$0.028) & 11.483 ($\pm$1.638) & 0.269 ($\pm$0.027) & 0.328 ($\pm$0.164) & 0.094 ($\pm$0.042) & 10.698 ($\pm$1.956) & 0.380 ($\pm$0.150)\\
&CRFILL & 0.602 ($\pm$0.309) & 0.566 ($\pm$0.236) & 2.825 ($\pm$1.764) & 0.038 ($\pm$0.106) & 0.211 ($\pm$0.119) & 0.183 ($\pm$0.091) & 7.918 ($\pm$2.238) & 0.655 ($\pm$0.118) & 0.549 ($\pm$0.266) & 0.556 ($\pm$0.213) & 2.872 ($\pm$1.697) & 0.053 ($\pm$0.108)\\
&GMCNN & 0.641 ($\pm$0.367) & 0.811 ($\pm$0.793) & 2.883 ($\pm$4.171) & -0.023 ($\pm$0.218) & 0.539 ($\pm$0.334) & 1.226 ($\pm$0.958) & 0.464 ($\pm$3.428) & 0.116 ($\pm$0.327) & 0.475 ($\pm$0.273) & 0.716 ($\pm$0.763) & 3.915 ($\pm$4.798) & 0.005 ($\pm$0.245)\\
&EDGE CONNECT & 0.411 ($\pm$0.249) & 0.094 ($\pm$0.052) & 10.882 ($\pm$2.293) & 0.259 ($\pm$0.165) & 0.145 ($\pm$0.060) & 0.112 ($\pm$0.048) & 9.928 ($\pm$1.945) & 0.212 ($\pm$0.056) & 0.267 ($\pm$0.147) & 0.085 ($\pm$0.049) & 11.350 ($\pm$2.386) & 0.374 ($\pm$0.137)\\
\bottomrule
\end{tabular}
\end{adjustbox}
\caption{Demonstration of a multi-model attack. Note that this produces just a single adversarial image, which all models subsequently inpaint. The results are after 500 iterations with a step size of $\frac{\epsilon}{50}$ are presented in the form $\mu\ (\pm\sigma$).}
\label{tab:all_model}
\end{table*}

\begin{table*}[t]
\begin{adjustbox}{scale=0.5,center}
\begin{tabular}{@{}llrrrr|rrrr|rrrr@{}}
\toprule
\multicolumn{2}{l}{Distance to:} &\multicolumn{4}{c|}{Original} & \multicolumn{4}{c|}{Adversarial Target} & \multicolumn{4}{c}{Benign} \\
$\epsilon$ & &Loss & $l_2$ & PSNR & SSIM & Loss & $l_2$ & PSNR & SSIM & Loss & $l_2$ & PSNR & SSIM\\
\midrule
\multirow{5}{*}{$ 0.0 $}
&GENERATIVE & 0.467 ($\pm$0.282) & 0.111 ($\pm$0.113) & 12.491 ($\pm$6.257) & 0.250 ($\pm$0.222) & 0.755 ($\pm$0.334) & 1.186 ($\pm$0.298) & -0.622 ($\pm$0.989) & 0.036 ($\pm$0.100) & 0.179 ($\pm$0.062) & 0.000 ($\pm$0.000) & 134.254 ($\pm$3.384) & 1.000 ($\pm$0.000)\\
&RFR & 0.279 ($\pm$0.220) & 0.027 ($\pm$0.034) & 19.156 ($\pm$7.224) & 0.387 ($\pm$0.261) & 0.433 ($\pm$0.191) & 0.292 ($\pm$0.072) & 5.464 ($\pm$0.992) & 0.090 ($\pm$0.055) & 0.001 ($\pm$0.001) & 0.000 ($\pm$0.000) & 144.880 ($\pm$3.990) & 1.000 ($\pm$0.000)\\
&RN & 0.300 ($\pm$0.227) & 0.025 ($\pm$0.023) & 17.891 ($\pm$4.161) & 0.420 ($\pm$0.220) & 0.473 ($\pm$0.204) & 0.292 ($\pm$0.060) & 5.438 ($\pm$0.851) & 0.104 ($\pm$0.051) & 0.001 ($\pm$0.001) & 0.000 ($\pm$0.000) & inf ($\pm$nan) & 1.000 ($\pm$0.000)\\
&CRFILL & 0.470 ($\pm$0.295) & 0.109 ($\pm$0.120) & 13.342 ($\pm$7.767) & 0.319 ($\pm$0.257) & 0.796 ($\pm$0.355) & 1.205 ($\pm$0.309) & -0.683 ($\pm$1.021) & 0.044 ($\pm$0.115) & 0.180 ($\pm$0.059) & 0.000 ($\pm$0.000) & inf ($\pm$nan) & 1.000 ($\pm$0.000)\\
&GMCNN & 0.485 ($\pm$0.267) & 0.122 ($\pm$0.108) & 10.891 ($\pm$4.560) & 0.210 ($\pm$0.208) & 0.721 ($\pm$0.289) & 1.136 ($\pm$0.283) & -0.432 ($\pm$1.014) & 0.047 ($\pm$0.110) & 0.181 ($\pm$0.060) & 0.000 ($\pm$0.000) & inf ($\pm$nan) & 1.000 ($\pm$0.000)\\
&EDGE CONNECT & 0.290 ($\pm$0.226) & 0.025 ($\pm$0.022) & 18.198 ($\pm$5.129) & 0.390 ($\pm$0.245) & 0.422 ($\pm$0.189) & 0.299 ($\pm$0.085) & 5.383 ($\pm$1.088) & 0.104 ($\pm$0.056) & 0.001 ($\pm$0.001) & 0.000 ($\pm$0.000) & 135.133 ($\pm$3.776) & 1.000 ($\pm$0.000)\\
\midrule
\multirow{5}{*}{$ 0.03 $}
&GENERATIVE & 0.438 ($\pm$0.248) & 0.106 ($\pm$0.113) & 12.386 ($\pm$5.484) & 0.236 ($\pm$0.212) & 0.643 ($\pm$0.253) & 1.125 ($\pm$0.269) & -0.406 ($\pm$0.938) & 0.044 ($\pm$0.098) & 0.273 ($\pm$0.121) & 0.018 ($\pm$0.015) & 19.479 ($\pm$4.839) & 0.569 ($\pm$0.123)\\
&RFR & 0.312 ($\pm$0.212) & 0.034 ($\pm$0.040) & 16.647 ($\pm$4.209) & 0.337 ($\pm$0.240) & 0.342 ($\pm$0.141) & 0.257 ($\pm$0.072) & 6.067 ($\pm$1.163) & 0.117 ($\pm$0.060) & 0.129 ($\pm$0.086) & 0.011 ($\pm$0.009) & 20.816 ($\pm$3.711) & 0.705 ($\pm$0.143)\\
&RN & 0.361 ($\pm$0.216) & 0.065 ($\pm$0.037) & 12.642 ($\pm$2.712) & 0.294 ($\pm$0.176) & 0.266 ($\pm$0.143) & 0.194 ($\pm$0.072) & 7.520 ($\pm$2.116) & 0.214 ($\pm$0.092) & 0.231 ($\pm$0.116) & 0.043 ($\pm$0.029) & 14.735 ($\pm$3.338) & 0.497 ($\pm$0.151)\\
&CRFILL & 0.453 ($\pm$0.244) & 0.146 ($\pm$0.161) & 9.908 ($\pm$3.587) & 0.205 ($\pm$0.174) & 0.603 ($\pm$0.260) & 1.001 ($\pm$0.284) & 0.165 ($\pm$1.236) & 0.076 ($\pm$0.105) & 0.344 ($\pm$0.137) & 0.072 ($\pm$0.061) & 12.690 ($\pm$3.306) & 0.405 ($\pm$0.151)\\
&GMCNN & 0.621 ($\pm$0.347) & 0.853 ($\pm$0.910) & 3.225 ($\pm$4.776) & -0.037 ($\pm$0.261) & 0.607 ($\pm$0.294) & 1.392 ($\pm$0.838) & -0.699 ($\pm$2.484) & 0.010 ($\pm$0.238) & 0.440 ($\pm$0.250) & 0.729 ($\pm$0.886) & 4.847 ($\pm$5.691) & 0.029 ($\pm$0.332)\\
&EDGE CONNECT & 0.334 ($\pm$0.213) & 0.053 ($\pm$0.030) & 13.325 ($\pm$2.230) & 0.320 ($\pm$0.201) & 0.253 ($\pm$0.114) & 0.233 ($\pm$0.079) & 6.547 ($\pm$1.323) & 0.193 ($\pm$0.058) & 0.159 ($\pm$0.087) & 0.036 ($\pm$0.027) & 15.380 ($\pm$2.813) & 0.603 ($\pm$0.146)\\
\midrule
\multirow{5}{*}{$ 0.05 $}
&GENERATIVE & 0.441 ($\pm$0.249) & 0.108 ($\pm$0.111) & 12.121 ($\pm$5.201) & 0.225 ($\pm$0.211) & 0.623 ($\pm$0.244) & 1.093 ($\pm$0.258) & -0.281 ($\pm$0.926) & 0.050 ($\pm$0.099) & 0.289 ($\pm$0.132) & 0.028 ($\pm$0.025) & 17.483 ($\pm$4.559) & 0.476 ($\pm$0.146)\\
&RFR & 0.332 ($\pm$0.219) & 0.040 ($\pm$0.041) & 15.533 ($\pm$3.600) & 0.325 ($\pm$0.228) & 0.310 ($\pm$0.132) & 0.232 ($\pm$0.073) & 6.547 ($\pm$1.356) & 0.139 ($\pm$0.063) & 0.161 ($\pm$0.100) & 0.019 ($\pm$0.013) & 18.374 ($\pm$3.171) & 0.630 ($\pm$0.149)\\
&RN & 0.386 ($\pm$0.225) & 0.081 ($\pm$0.047) & 11.697 ($\pm$2.776) & 0.266 ($\pm$0.163) & 0.203 ($\pm$0.115) & 0.153 ($\pm$0.071) & 8.763 ($\pm$2.617) & 0.239 ($\pm$0.098) & 0.271 ($\pm$0.133) & 0.062 ($\pm$0.040) & 13.094 ($\pm$3.125) & 0.435 ($\pm$0.150)\\
&CRFILL & 0.491 ($\pm$0.241) & 0.244 ($\pm$0.229) & 7.291 ($\pm$3.089) & 0.128 ($\pm$0.130) & 0.509 ($\pm$0.254) & 0.782 ($\pm$0.313) & 1.481 ($\pm$2.070) & 0.135 ($\pm$0.147) & 0.406 ($\pm$0.156) & 0.186 ($\pm$0.167) & 8.676 ($\pm$3.444) & 0.232 ($\pm$0.141)\\
&GMCNN & 0.605 ($\pm$0.329) & 0.818 ($\pm$0.824) & 3.047 ($\pm$4.426) & -0.018 ($\pm$0.246) & 0.579 ($\pm$0.294) & 1.349 ($\pm$0.869) & -0.453 ($\pm$2.674) & 0.040 ($\pm$0.259) & 0.432 ($\pm$0.234) & 0.698 ($\pm$0.801) & 4.491 ($\pm$5.206) & 0.038 ($\pm$0.301)\\
&EDGE CONNECT & 0.351 ($\pm$0.219) & 0.057 ($\pm$0.033) & 13.032 ($\pm$2.216) & 0.317 ($\pm$0.200) & 0.216 ($\pm$0.100) & 0.206 ($\pm$0.078) & 7.120 ($\pm$1.481) & 0.214 ($\pm$0.056) & 0.190 ($\pm$0.102) & 0.042 ($\pm$0.031) & 14.655 ($\pm$2.658) & 0.546 ($\pm$0.151)\\
\midrule
\multirow{5}{*}{$ 0.1 $}
&GENERATIVE & 0.452 ($\pm$0.255) & 0.119 ($\pm$0.123) & 11.390 ($\pm$4.767) & 0.202 ($\pm$0.209) & 0.585 ($\pm$0.228) & 1.015 ($\pm$0.236) & 0.035 ($\pm$0.917) & 0.068 ($\pm$0.102) & 0.317 ($\pm$0.150) & 0.050 ($\pm$0.044) & 14.609 ($\pm$3.986) & 0.346 ($\pm$0.169)\\
&RFR & 0.365 ($\pm$0.229) & 0.055 ($\pm$0.045) & 13.639 ($\pm$2.979) & 0.314 ($\pm$0.211) & 0.244 ($\pm$0.115) & 0.184 ($\pm$0.075) & 7.698 ($\pm$1.801) & 0.190 ($\pm$0.066) & 0.213 ($\pm$0.118) & 0.037 ($\pm$0.022) & 15.165 ($\pm$2.771) & 0.504 ($\pm$0.172)\\
&RN & 0.420 ($\pm$0.238) & 0.110 ($\pm$0.065) & 10.333 ($\pm$2.615) & 0.234 ($\pm$0.144) & 0.133 ($\pm$0.084) & 0.102 ($\pm$0.061) & 10.819 ($\pm$3.101) & 0.278 ($\pm$0.098) & 0.319 ($\pm$0.153) & 0.092 ($\pm$0.054) & 11.133 ($\pm$2.737) & 0.368 ($\pm$0.140)\\
&CRFILL & 0.581 ($\pm$0.264) & 0.508 ($\pm$0.299) & 3.579 ($\pm$2.371) & 0.054 ($\pm$0.094) & 0.316 ($\pm$0.213) & 0.366 ($\pm$0.306) & 5.892 ($\pm$3.956) & 0.349 ($\pm$0.246) & 0.518 ($\pm$0.208) & 0.478 ($\pm$0.273) & 3.861 ($\pm$2.430) & 0.079 ($\pm$0.100)\\
&GMCNN & 0.650 ($\pm$0.372) & 0.953 ($\pm$0.896) & 2.254 ($\pm$4.355) & -0.049 ($\pm$0.243) & 0.570 ($\pm$0.329) & 1.379 ($\pm$0.954) & -0.283 ($\pm$3.164) & 0.055 ($\pm$0.303) & 0.476 ($\pm$0.273) & 0.825 ($\pm$0.860) & 3.409 ($\pm$4.919) & -0.024 ($\pm$0.285)\\
&EDGE CONNECT & 0.382 ($\pm$0.231) & 0.072 ($\pm$0.051) & 12.158 ($\pm$2.498) & 0.307 ($\pm$0.193) & 0.158 ($\pm$0.075) & 0.158 ($\pm$0.074) & 8.411 ($\pm$1.845) & 0.238 ($\pm$0.050) & 0.233 ($\pm$0.126) & 0.060 ($\pm$0.049) & 13.205 ($\pm$2.890) & 0.476 ($\pm$0.153)\\
\midrule
\multirow{5}{*}{$ 0.2 $}
&GENERATIVE & 0.473 ($\pm$0.265) & 0.150 ($\pm$0.145) & 9.902 ($\pm$3.962) & 0.163 ($\pm$0.196) & 0.519 ($\pm$0.205) & 0.879 ($\pm$0.214) & 0.674 ($\pm$0.967) & 0.104 ($\pm$0.107) & 0.355 ($\pm$0.173) & 0.093 ($\pm$0.078) & 11.588 ($\pm$3.386) & 0.226 ($\pm$0.173)\\
&RFR & 0.395 ($\pm$0.238) & 0.081 ($\pm$0.064) & 11.838 ($\pm$2.770) & 0.301 ($\pm$0.199) & 0.172 ($\pm$0.093) & 0.135 ($\pm$0.066) & 9.268 ($\pm$2.317) & 0.232 ($\pm$0.052) & 0.258 ($\pm$0.134) & 0.064 ($\pm$0.037) & 12.670 ($\pm$2.637) & 0.414 ($\pm$0.181)\\
&RN & 0.445 ($\pm$0.246) & 0.140 ($\pm$0.075) & 9.130 ($\pm$2.277) & 0.222 ($\pm$0.142) & 0.085 ($\pm$0.057) & 0.063 ($\pm$0.044) & 13.022 ($\pm$3.056) & 0.308 ($\pm$0.083) & 0.351 ($\pm$0.164) & 0.124 ($\pm$0.061) & 9.574 ($\pm$2.225) & 0.327 ($\pm$0.130)\\
&CRFILL & 0.671 ($\pm$0.335) & 0.801 ($\pm$0.338) & 1.328 ($\pm$1.803) & 0.040 ($\pm$0.098) & 0.185 ($\pm$0.116) & 0.104 ($\pm$0.107) & 11.718 ($\pm$4.190) & 0.598 ($\pm$0.231) & 0.618 ($\pm$0.294) & 0.788 ($\pm$0.308) & 1.368 ($\pm$1.730) & 0.053 ($\pm$0.101)\\
&GMCNN & 0.663 ($\pm$0.361) & 1.035 ($\pm$0.892) & 1.610 ($\pm$4.047) & -0.067 ($\pm$0.246) & 0.542 ($\pm$0.341) & 1.396 ($\pm$1.043) & 0.053 ($\pm$3.807) & 0.078 ($\pm$0.356) & 0.492 ($\pm$0.273) & 0.916 ($\pm$0.882) & 2.670 ($\pm$4.723) & -0.055 ($\pm$0.268)\\
&EDGE CONNECT & 0.414 ($\pm$0.249) & 0.096 ($\pm$0.065) & 11.004 ($\pm$2.691) & 0.294 ($\pm$0.190) & 0.107 ($\pm$0.051) & 0.111 ($\pm$0.064) & 10.150 ($\pm$2.228) & 0.259 ($\pm$0.045) & 0.273 ($\pm$0.149) & 0.086 ($\pm$0.063) & 11.646 ($\pm$3.006) & 0.421 ($\pm$0.166)\\
\midrule
\multirow{5}{*}{$ 0.3 $}
&GENERATIVE & 0.489 ($\pm$0.271) & 0.184 ($\pm$0.165) & 8.676 ($\pm$3.404) & 0.132 ($\pm$0.178) & 0.466 ($\pm$0.189) & 0.768 ($\pm$0.208) & 1.284 ($\pm$1.082) & 0.140 ($\pm$0.114) & 0.379 ($\pm$0.187) & 0.135 ($\pm$0.103) & 9.746 ($\pm$2.973) & 0.167 ($\pm$0.163)\\
&RFR & 0.412 ($\pm$0.242) & 0.095 ($\pm$0.064) & 10.965 ($\pm$2.496) & 0.288 ($\pm$0.188) & 0.143 ($\pm$0.078) & 0.110 ($\pm$0.055) & 10.191 ($\pm$2.460) & 0.247 ($\pm$0.043) & 0.278 ($\pm$0.140) & 0.079 ($\pm$0.041) & 11.604 ($\pm$2.382) & 0.381 ($\pm$0.173)\\
&RN & 0.456 ($\pm$0.251) & 0.154 ($\pm$0.076) & 8.620 ($\pm$2.073) & 0.216 ($\pm$0.137) & 0.067 ($\pm$0.045) & 0.049 ($\pm$0.038) & 14.125 ($\pm$3.057) & 0.322 ($\pm$0.086) & 0.365 ($\pm$0.169) & 0.140 ($\pm$0.062) & 8.970 ($\pm$1.995) & 0.315 ($\pm$0.129)\\
&CRFILL & 0.698 ($\pm$0.352) & 0.896 ($\pm$0.340) & 0.759 ($\pm$1.550) & 0.035 ($\pm$0.100) & 0.159 ($\pm$0.094) & 0.060 ($\pm$0.058) & 14.076 ($\pm$4.019) & 0.695 ($\pm$0.184) & 0.646 ($\pm$0.311) & 0.884 ($\pm$0.309) & 0.789 ($\pm$1.480) & 0.047 ($\pm$0.100)\\
&GMCNN & 0.681 ($\pm$0.383) & 1.112 ($\pm$0.849) & 0.968 ($\pm$3.710) & -0.067 ($\pm$0.239) & 0.533 ($\pm$0.366) & 1.457 ($\pm$1.103) & 0.175 ($\pm$4.346) & 0.081 ($\pm$0.395) & 0.511 ($\pm$0.284) & 0.993 ($\pm$0.843) & 1.851 ($\pm$4.251) & -0.066 ($\pm$0.258)\\
&EDGE CONNECT & 0.428 ($\pm$0.256) & 0.112 ($\pm$0.075) & 10.336 ($\pm$2.672) & 0.285 ($\pm$0.186) & 0.089 ($\pm$0.042) & 0.091 ($\pm$0.057) & 11.166 ($\pm$2.523) & 0.268 ($\pm$0.047) & 0.289 ($\pm$0.158) & 0.103 ($\pm$0.073) & 10.865 ($\pm$3.012) & 0.398 ($\pm$0.168)\\
\bottomrule
\end{tabular}
\end{adjustbox}
\caption{Impact of markpainting attack on each model individually: the model in each row is attacked and the results presented are from evaluation on that same model. This table uses the same input/target/mask combinations as~\Cref{tab:single_model}. The results are after 100 iterations with a step size of $\frac{\epsilon}{50}$ and are presented in the form $\mu\ (\pm\sigma$).}
\label{tab:individual_model}
\end{table*}

\begin{table*}[t]
\begin{adjustbox}{scale=0.5,center}
\begin{tabular}{@{}llrrrr|rrrr|rrrr@{}}
\toprule
\multicolumn{2}{l}{Distance to:} &\multicolumn{4}{c|}{Original} & \multicolumn{4}{c|}{Adversarial Target} & \multicolumn{4}{c}{Benign} \\
$\epsilon$ & &Loss & $l_2$ & PSNR & SSIM & Loss & $l_2$ & PSNR & SSIM & Loss & $l_2$ & PSNR & SSIM\\
\midrule
\multirow{5}{*}{$ 0.0 $}
&GENERATIVE & 0.467 ($\pm$0.282) & 0.111 ($\pm$0.113) & 12.491 ($\pm$6.257) & 0.250 ($\pm$0.222) & 0.755 ($\pm$0.334) & 1.186 ($\pm$0.298) & -0.622 ($\pm$0.989) & 0.036 ($\pm$0.100) & 0.179 ($\pm$0.062) & 0.000 ($\pm$0.000) & 134.217 ($\pm$3.413) & 1.000 ($\pm$0.000)\\
&RFR & 0.279 ($\pm$0.220) & 0.027 ($\pm$0.034) & 19.156 ($\pm$7.224) & 0.387 ($\pm$0.261) & 0.433 ($\pm$0.191) & 0.292 ($\pm$0.072) & 5.464 ($\pm$0.992) & 0.090 ($\pm$0.055) & 0.001 ($\pm$0.001) & 0.000 ($\pm$0.000) & 144.923 ($\pm$3.966) & 1.000 ($\pm$0.000)\\
&RN & 0.300 ($\pm$0.227) & 0.025 ($\pm$0.023) & 17.891 ($\pm$4.161) & 0.420 ($\pm$0.220) & 0.473 ($\pm$0.204) & 0.292 ($\pm$0.060) & 5.438 ($\pm$0.851) & 0.104 ($\pm$0.051) & 0.001 ($\pm$0.001) & 0.000 ($\pm$0.000) & inf ($\pm$nan) & 1.000 ($\pm$0.000)\\
&CRFILL & 0.470 ($\pm$0.295) & 0.109 ($\pm$0.120) & 13.342 ($\pm$7.767) & 0.319 ($\pm$0.257) & 0.796 ($\pm$0.355) & 1.205 ($\pm$0.309) & -0.683 ($\pm$1.021) & 0.044 ($\pm$0.115) & 0.180 ($\pm$0.059) & 0.000 ($\pm$0.000) & inf ($\pm$nan) & 1.000 ($\pm$0.000)\\
&GMCNN & 0.485 ($\pm$0.267) & 0.122 ($\pm$0.108) & 10.891 ($\pm$4.560) & 0.210 ($\pm$0.208) & 0.721 ($\pm$0.289) & 1.136 ($\pm$0.283) & -0.432 ($\pm$1.014) & 0.047 ($\pm$0.110) & 0.181 ($\pm$0.060) & 0.000 ($\pm$0.000) & inf ($\pm$nan) & 1.000 ($\pm$0.000)\\
&EDGE CONNECT & 0.290 ($\pm$0.226) & 0.025 ($\pm$0.022) & 18.198 ($\pm$5.129) & 0.390 ($\pm$0.245) & 0.422 ($\pm$0.189) & 0.299 ($\pm$0.085) & 5.383 ($\pm$1.088) & 0.104 ($\pm$0.056) & 0.001 ($\pm$0.001) & 0.000 ($\pm$0.000) & 135.041 ($\pm$3.839) & 1.000 ($\pm$0.000)\\
\midrule
\multirow{5}{*}{$ 0.03 $}
&GENERATIVE & 0.466 ($\pm$0.280) & 0.110 ($\pm$0.113) & 12.551 ($\pm$6.370) & 0.245 ($\pm$0.219) & 0.748 ($\pm$0.330) & 1.175 ($\pm$0.298) & -0.579 ($\pm$0.997) & 0.036 ($\pm$0.098) & 0.214 ($\pm$0.085) & 0.005 ($\pm$0.006) & 26.037 ($\pm$6.145) & 0.829 ($\pm$0.100)\\
&RFR & 0.287 ($\pm$0.219) & 0.028 ($\pm$0.035) & 18.593 ($\pm$6.047) & 0.362 ($\pm$0.252) & 0.416 ($\pm$0.184) & 0.282 ($\pm$0.071) & 5.617 ($\pm$1.011) & 0.088 ($\pm$0.053) & 0.032 ($\pm$0.027) & 0.001 ($\pm$0.001) & 30.462 ($\pm$3.392) & 0.948 ($\pm$0.038)\\
&RN & 0.361 ($\pm$0.216) & 0.065 ($\pm$0.037) & 12.642 ($\pm$2.712) & 0.294 ($\pm$0.176) & 0.266 ($\pm$0.143) & 0.194 ($\pm$0.072) & 7.520 ($\pm$2.116) & 0.214 ($\pm$0.092) & 0.231 ($\pm$0.116) & 0.043 ($\pm$0.029) & 14.735 ($\pm$3.338) & 0.497 ($\pm$0.151)\\
&CRFILL & 0.472 ($\pm$0.296) & 0.107 ($\pm$0.115) & 13.200 ($\pm$7.266) & 0.299 ($\pm$0.243) & 0.786 ($\pm$0.353) & 1.180 ($\pm$0.308) & -0.589 ($\pm$1.036) & 0.044 ($\pm$0.109) & 0.230 ($\pm$0.090) & 0.008 ($\pm$0.008) & 23.472 ($\pm$5.608) & 0.809 ($\pm$0.092)\\
&GMCNN & 0.484 ($\pm$0.266) & 0.121 ($\pm$0.103) & 10.816 ($\pm$4.359) & 0.206 ($\pm$0.201) & 0.717 ($\pm$0.289) & 1.129 ($\pm$0.282) & -0.405 ($\pm$1.019) & 0.048 ($\pm$0.108) & 0.195 ($\pm$0.074) & 0.005 ($\pm$0.019) & 31.205 ($\pm$5.169) & 0.922 ($\pm$0.071)\\
&EDGE CONNECT & 0.291 ($\pm$0.223) & 0.025 ($\pm$0.021) & 18.041 ($\pm$4.864) & 0.378 ($\pm$0.239) & 0.407 ($\pm$0.180) & 0.293 ($\pm$0.088) & 5.489 ($\pm$1.139) & 0.107 ($\pm$0.056) & 0.036 ($\pm$0.030) & 0.001 ($\pm$0.001) & 29.890 ($\pm$3.880) & 0.911 ($\pm$0.049)\\
\midrule
\multirow{5}{*}{$ 0.05 $}
&GENERATIVE & 0.466 ($\pm$0.279) & 0.110 ($\pm$0.110) & 12.552 ($\pm$6.336) & 0.240 ($\pm$0.215) & 0.745 ($\pm$0.329) & 1.167 ($\pm$0.299) & -0.548 ($\pm$1.006) & 0.035 ($\pm$0.098) & 0.228 ($\pm$0.096) & 0.009 ($\pm$0.011) & 23.663 ($\pm$6.018) & 0.760 ($\pm$0.127)\\
&RFR & 0.296 ($\pm$0.220) & 0.028 ($\pm$0.035) & 18.146 ($\pm$5.525) & 0.340 ($\pm$0.241) & 0.408 ($\pm$0.180) & 0.273 ($\pm$0.070) & 5.761 ($\pm$1.029) & 0.086 ($\pm$0.052) & 0.059 ($\pm$0.051) & 0.003 ($\pm$0.003) & 26.769 ($\pm$3.464) & 0.899 ($\pm$0.070)\\
&RN & 0.386 ($\pm$0.225) & 0.081 ($\pm$0.047) & 11.697 ($\pm$2.776) & 0.266 ($\pm$0.163) & 0.203 ($\pm$0.115) & 0.153 ($\pm$0.071) & 8.763 ($\pm$2.617) & 0.239 ($\pm$0.098) & 0.271 ($\pm$0.133) & 0.062 ($\pm$0.040) & 13.094 ($\pm$3.125) & 0.435 ($\pm$0.150)\\
&CRFILL & 0.478 ($\pm$0.300) & 0.108 ($\pm$0.116) & 12.947 ($\pm$6.918) & 0.281 ($\pm$0.235) & 0.782 ($\pm$0.356) & 1.157 ($\pm$0.305) & -0.503 ($\pm$1.040) & 0.045 ($\pm$0.105) & 0.266 ($\pm$0.124) & 0.016 ($\pm$0.016) & 20.294 ($\pm$5.380) & 0.703 ($\pm$0.123)\\
&GMCNN & 0.485 ($\pm$0.265) & 0.121 ($\pm$0.101) & 10.789 ($\pm$4.306) & 0.204 ($\pm$0.198) & 0.717 ($\pm$0.289) & 1.126 ($\pm$0.282) & -0.392 ($\pm$1.022) & 0.049 ($\pm$0.108) & 0.198 ($\pm$0.071) & 0.005 ($\pm$0.018) & 28.754 ($\pm$4.657) & 0.888 ($\pm$0.079)\\
&EDGE CONNECT & 0.293 ($\pm$0.221) & 0.025 ($\pm$0.021) & 17.837 ($\pm$4.701) & 0.368 ($\pm$0.235) & 0.400 ($\pm$0.175) & 0.289 ($\pm$0.090) & 5.563 ($\pm$1.173) & 0.107 ($\pm$0.056) & 0.054 ($\pm$0.044) & 0.003 ($\pm$0.002) & 27.208 ($\pm$3.807) & 0.860 ($\pm$0.068)\\
\midrule
\multirow{5}{*}{$ 0.1 $}
&GENERATIVE & 0.469 ($\pm$0.278) & 0.111 ($\pm$0.114) & 12.406 ($\pm$6.074) & 0.228 ($\pm$0.202) & 0.739 ($\pm$0.323) & 1.149 ($\pm$0.300) & -0.475 ($\pm$1.020) & 0.034 ($\pm$0.092) & 0.254 ($\pm$0.118) & 0.015 ($\pm$0.015) & 20.624 ($\pm$5.403) & 0.642 ($\pm$0.151)\\
&RFR & 0.318 ($\pm$0.224) & 0.032 ($\pm$0.038) & 17.145 ($\pm$4.658) & 0.299 ($\pm$0.216) & 0.394 ($\pm$0.169) & 0.254 ($\pm$0.070) & 6.103 ($\pm$1.099) & 0.082 ($\pm$0.048) & 0.107 ($\pm$0.078) & 0.008 ($\pm$0.007) & 22.348 ($\pm$3.283) & 0.801 ($\pm$0.102)\\
&RN & 0.420 ($\pm$0.238) & 0.110 ($\pm$0.065) & 10.333 ($\pm$2.615) & 0.234 ($\pm$0.144) & 0.133 ($\pm$0.084) & 0.102 ($\pm$0.061) & 10.819 ($\pm$3.101) & 0.278 ($\pm$0.098) & 0.319 ($\pm$0.153) & 0.092 ($\pm$0.054) & 11.133 ($\pm$2.737) & 0.368 ($\pm$0.140)\\
&CRFILL & 0.499 ($\pm$0.315) & 0.117 ($\pm$0.124) & 12.039 ($\pm$5.854) & 0.240 ($\pm$0.216) & 0.770 ($\pm$0.357) & 1.091 ($\pm$0.292) & -0.244 ($\pm$1.048) & 0.052 ($\pm$0.095) & 0.334 ($\pm$0.181) & 0.038 ($\pm$0.037) & 16.156 ($\pm$4.685) & 0.529 ($\pm$0.164)\\
&GMCNN & 0.485 ($\pm$0.265) & 0.121 ($\pm$0.097) & 10.729 ($\pm$4.167) & 0.200 ($\pm$0.191) & 0.714 ($\pm$0.287) & 1.118 ($\pm$0.281) & -0.359 ($\pm$1.029) & 0.049 ($\pm$0.106) & 0.207 ($\pm$0.078) & 0.007 ($\pm$0.023) & 25.782 ($\pm$4.223) & 0.824 ($\pm$0.099)\\
&EDGE CONNECT & 0.304 ($\pm$0.222) & 0.028 ($\pm$0.021) & 17.147 ($\pm$4.125) & 0.344 ($\pm$0.222) & 0.392 ($\pm$0.169) & 0.280 ($\pm$0.092) & 5.714 ($\pm$1.234) & 0.106 ($\pm$0.054) & 0.090 ($\pm$0.067) & 0.006 ($\pm$0.005) & 23.328 ($\pm$3.647) & 0.759 ($\pm$0.099)\\
\midrule
\multirow{5}{*}{$ 0.2 $}
&GENERATIVE & 0.478 ($\pm$0.280) & 0.113 ($\pm$0.110) & 12.037 ($\pm$5.618) & 0.204 ($\pm$0.183) & 0.734 ($\pm$0.322) & 1.119 ($\pm$0.302) & -0.355 ($\pm$1.048) & 0.032 ($\pm$0.084) & 0.285 ($\pm$0.135) & 0.027 ($\pm$0.023) & 17.797 ($\pm$4.840) & 0.502 ($\pm$0.166)\\
&RFR & 0.352 ($\pm$0.234) & 0.039 ($\pm$0.043) & 15.691 ($\pm$3.755) & 0.245 ($\pm$0.172) & 0.382 ($\pm$0.160) & 0.227 ($\pm$0.071) & 6.632 ($\pm$1.255) & 0.078 ($\pm$0.043) & 0.161 ($\pm$0.108) & 0.016 ($\pm$0.014) & 18.952 ($\pm$3.007) & 0.692 ($\pm$0.120)\\
&RN & 0.445 ($\pm$0.246) & 0.140 ($\pm$0.075) & 9.130 ($\pm$2.277) & 0.222 ($\pm$0.142) & 0.085 ($\pm$0.057) & 0.063 ($\pm$0.044) & 13.022 ($\pm$3.056) & 0.308 ($\pm$0.083) & 0.351 ($\pm$0.164) & 0.124 ($\pm$0.061) & 9.574 ($\pm$2.225) & 0.327 ($\pm$0.130)\\
&CRFILL & 0.526 ($\pm$0.321) & 0.139 ($\pm$0.136) & 10.363 ($\pm$4.237) & 0.181 ($\pm$0.177) & 0.736 ($\pm$0.333) & 0.973 ($\pm$0.273) & 0.265 ($\pm$1.077) & 0.061 ($\pm$0.083) & 0.405 ($\pm$0.227) & 0.078 ($\pm$0.065) & 12.485 ($\pm$3.724) & 0.347 ($\pm$0.167)\\
&GMCNN & 0.489 ($\pm$0.266) & 0.122 ($\pm$0.091) & 10.537 ($\pm$3.947) & 0.189 ($\pm$0.176) & 0.713 ($\pm$0.288) & 1.104 ($\pm$0.280) & -0.301 ($\pm$1.041) & 0.049 ($\pm$0.101) & 0.225 ($\pm$0.088) & 0.013 ($\pm$0.035) & 22.891 ($\pm$4.258) & 0.738 ($\pm$0.122)\\
&EDGE CONNECT & 0.322 ($\pm$0.225) & 0.032 ($\pm$0.023) & 16.066 ($\pm$3.440) & 0.307 ($\pm$0.194) & 0.380 ($\pm$0.161) & 0.268 ($\pm$0.093) & 5.937 ($\pm$1.300) & 0.107 ($\pm$0.052) & 0.131 ($\pm$0.088) & 0.013 ($\pm$0.010) & 20.012 ($\pm$3.443) & 0.645 ($\pm$0.117)\\
\midrule
\multirow{5}{*}{$ 0.3 $}
&GENERATIVE & 0.487 ($\pm$0.277) & 0.115 ($\pm$0.109) & 11.625 ($\pm$5.054) & 0.169 ($\pm$0.144) & 0.734 ($\pm$0.317) & 1.099 ($\pm$0.303) & -0.270 ($\pm$1.068) & 0.025 ($\pm$0.068) & 0.312 ($\pm$0.149) & 0.036 ($\pm$0.028) & 16.181 ($\pm$4.334) & 0.400 ($\pm$0.152)\\
&RFR & 0.373 ($\pm$0.237) & 0.043 ($\pm$0.038) & 14.869 ($\pm$3.380) & 0.215 ($\pm$0.147) & 0.374 ($\pm$0.153) & 0.211 ($\pm$0.072) & 6.988 ($\pm$1.393) & 0.076 ($\pm$0.039) & 0.191 ($\pm$0.116) & 0.022 ($\pm$0.015) & 17.416 ($\pm$2.858) & 0.633 ($\pm$0.123)\\
&RN & 0.456 ($\pm$0.251) & 0.154 ($\pm$0.076) & 8.620 ($\pm$2.073) & 0.216 ($\pm$0.137) & 0.067 ($\pm$0.045) & 0.049 ($\pm$0.038) & 14.125 ($\pm$3.057) & 0.322 ($\pm$0.086) & 0.365 ($\pm$0.169) & 0.140 ($\pm$0.062) & 8.970 ($\pm$1.995) & 0.315 ($\pm$0.129)\\
&CRFILL & 0.541 ($\pm$0.320) & 0.157 ($\pm$0.126) & 9.376 ($\pm$3.685) & 0.133 ($\pm$0.127) & 0.710 ($\pm$0.322) & 0.895 ($\pm$0.267) & 0.642 ($\pm$1.136) & 0.063 ($\pm$0.076) & 0.439 ($\pm$0.239) & 0.106 ($\pm$0.071) & 10.830 ($\pm$3.355) & 0.247 ($\pm$0.132)\\
&GMCNN & 0.493 ($\pm$0.266) & 0.127 ($\pm$0.101) & 10.359 ($\pm$3.862) & 0.173 ($\pm$0.160) & 0.712 ($\pm$0.286) & 1.095 ($\pm$0.290) & -0.257 ($\pm$1.078) & 0.046 ($\pm$0.095) & 0.236 ($\pm$0.090) & 0.014 ($\pm$0.030) & 21.002 ($\pm$3.783) & 0.664 ($\pm$0.120)\\
&EDGE CONNECT & 0.334 ($\pm$0.225) & 0.036 ($\pm$0.024) & 15.416 ($\pm$3.074) & 0.282 ($\pm$0.180) & 0.373 ($\pm$0.157) & 0.259 ($\pm$0.093) & 6.092 ($\pm$1.351) & 0.106 ($\pm$0.051) & 0.153 ($\pm$0.098) & 0.018 ($\pm$0.014) & 18.638 ($\pm$3.443) & 0.586 ($\pm$0.124)\\
\bottomrule
\end{tabular}
\end{adjustbox}
\caption{Key attribution based on the adversarial sample produced. RN was attacked. The results are after 100 iterations with a step size of $\frac{\epsilon}{50}$.  The results are presented in the form $\mu\ (\pm\sigma$)}
\label{tab:single_model}
\end{table*}

\end{document}